\documentclass[10pt,twocolumn,letterpaper]{article}

\usepackage[pagenumbers]{iccv} 

\definecolor{iccvblue}{rgb}{0.21,0.49,0.74}
\usepackage[pagebackref,breaklinks,colorlinks,allcolors=iccvblue]{hyperref}
\usepackage{bbding}
\usepackage{multirow}
\usepackage {arydshln}
\usepackage{float} 
\usepackage{pifont}

\usepackage{marvosym}

\def\paperID{9284} 
\def\confName{ICCV}
\def\confYear{2025}

\title{Token Activation Map to Visually Explain Multimodal LLMs}

\author{Yi Li, Hualiang Wang, Xinpeng Ding, Haonan Wang, Xiaomeng Li\Letter \\
\small Department of Electronic and Computer Engineering, The Hong Kong University of Science and Technology, Hong Kong SAR, China. \\
\texttt\small{\{ylini,xmli\}@ust.hk} \\
}

\begin{document}
\maketitle
\begin{abstract}
Multimodal large language models (MLLMs) are broadly empowering various fields. Despite their advancements, the explainability of MLLMs remains less explored, hindering deeper understanding, model credibility, and effective visualization. Unlike conventional vision models (e.g., CNNs, ViTs, CLIP) that produce a single output, MLLMs generate sequences of tokens progressively, where each generated token depends on the previous context. Therefore, earlier context tokens can introduce redundant activations that interfere with the explanation of later tokens beyond their original information. Existing studies often overlook this issue, but our observations reveal that these redundant correlations can significantly hurt the reliability of explanations. To address this, we propose an estimated causal inference method to mitigate the interference of context to achieve high-quality MLLM explanation, with a novel rank Gaussian filter to further reduce activation noises. We term this method Token Activation Map (TAM) to highlight the consideration of interactions between tokens. TAM also indicates that it excels at explaining multiple tokens of MLLM, which is different from the Class Activation Map (CAM) for a single prediction. Our TAM method significantly outperforms existing SoTA methods, showcasing high-quality visualization results that can be utilized for various scenarios, such as object localization, failure case analysis, video visualization, MLLMs visual comparison, and model understanding (e.g., color, shape, action, location, visual reasoning, multi-turn conversation, etc). The code is available at \href{github.com/xmed-lab/TAM}{github.com/xmed-lab/TAM}.
\end{abstract}

\section{Introduction}
\label{sec:intro}

\begin{figure}[t]
\centering
 \includegraphics[width=8.4cm]{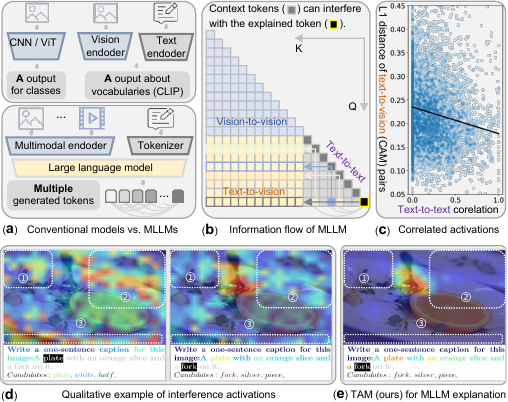}
\caption{Illustration of Motivation. (\textbf{a}) MLLMs generate multiple tokens beyond a single output. (\textbf{b}) The information flow indicates that MLLM generates tokens progressively, where one generated token (each row) is correlated with the context (prompt + earlier answer tokens). (\textbf{c}) We randomly pair CAMs and count their L1 distance against text correlation. Higher text correlation corresponds to lower distance, indicating concurrent interferences. (\textbf{d}) The example shows that the context token “plate” introduces interference activations (marked by white boxes) to the later explained token “fork”. (\textbf{e}) Our TAM reveals the original information of the explained token “fork”, eliminating context interferences and beyond CAM for a single output in (d). Results in (c, d, e) derive from Qwen2-VL-2B on the COCO Caption dataset.}
\label{fig_intro}
\end{figure}

Multimodal large language models (MLLMs or Multimodal LLMs) increasingly empower wide applications \cite{liu2023medical,lee2024llava,li2024manipllm,cui2024survey}, enabling multimodal inputs (e.g., images, videos, text) and human-like conversations. 
Although extensive efforts have been devoted to enhancing the performance of MLLMs \cite{alayrac2022flamingo,chen2024internvl,team2023gemini,liu2024visual,wang2024qwen2}, research on MLLM explainability remains less explored. It is crucial for user trust, model understanding, case analysis, and visualization. However, exploring the explainability of MLLMs is more challenging than conventional visual models (such as CNN \cite{he2016deep}, ViT \cite{dosovitskiy2020image}, CLIP \cite{radford2021learning}), where MLLMs progressively generate multiple tokens beyond a single output as Fig. \ref{fig_intro}a. Thus, we aim to explain all tokens of MLLM, revealing the visual cues via token-level activation maps (e.g., Fig. \ref{fig_intro}e).

Current MLLM explanation works \cite{zhang2024redundancy,teyunderstanding,ben2024lvlm} are generally borrowed from techniques \cite{selvaraju2017grad,omeiza2019smooth} for conventional models. For example, LLaVA-CAM \cite{zhang2024redundancy} uses Smooth Grad-CAM++ \cite{omeiza2019smooth} to explain LLaVA \cite{liu2024visual}. The self-attention weights \cite{vaswani2017attention} are applied to understand math problem-solving \cite{teyunderstanding}. Besides, LVLM-Interpret \cite{ben2024lvlm} integrates some conventional methods as a tool. Unfortunately, existing visual explainability methods are designed for conventional models \cite{he2016deep,dosovitskiy2020image,radford2021learning} that produce a single output, such as class activation maps (CAM) series \cite{zhou2016learning,selvaraju2017grad,chattopadhay2018grad,jiang2021layercam}, attention-based approaches \cite{lapuschkin2019unmasking,abnar2020quantifying}, attention with relevance \cite{achtibat2024attnlrp,chefer2021transformer}, model-agnostic methods \cite{ribeiro2016should,lundberg2017unified}, or multimodal explanation \cite{li2025closer,aflalo2022vl,li2022exploring,darcetvision}. \emph{However, these methods overlook that MLLMs generate multiple tokens progressively, where earlier context tokens interfere with later tokens to be explained}. As shown in Fig. \ref{fig_intro}d, the conventional CAM \cite{zhou2016learning} for the token  ``fork" is confounded by numerous redundant activations concurrent with the token ``plate" in context (context refers to prompt and earlier answer tokens than the explained token). Quantitative analysis in Fig. \ref{fig_intro}c reveals that this issue is widespread, where correlated tokens lead to close and interference activations. These interferences obscure the original information conveyed by the explained token, significantly degrading the explainability of MLLMs. Notably, this activation interference across tokens is unique to MLLMs and has been overlooked.

We aim to reveal the original information for the explained token while minimizing interference from context, as discussed above. Herein, the original information reveals a causal relation \cite{pearl2009causality} between the current token and the input, while the interferences represent the correlations that need to be excluded. Inspired by the potential outcome model (POM) of causal inference \cite{jiao2024causal}, we propose the estimated causal inference to explore this causal relation. Unlike raw causal inference by comparing ready-made results (yes vs. no), there is no single given output to represent the “no” result. Therefore, we estimated interferences from multiple context tokens that are highly related to the target, with a scale factor optimized by the least-squares method to generate the refined causal activation map. We also propose a novel rank Gaussian filter to reduce noises, thereby further enhancing the quality of activation maps. The above methods are collectively referred to as token activation map (TAM), emphasizing the consideration of interactions between tokens. TAM also indicates that it excels at explaining tokens in multiple rounds of MLLM, which differs from the Class Activation Map (CAM) for a single prediction. 

Experimentally, the proposed TAM demonstrates quantitative improvement over existing SoTA explainability methods by more than 8.96\% on the COCO Caption dataset \cite{chen2015microsoft} and 8.54\% on the OpenPSG dataset \cite{zhou2024openpsg} when applied to visually explain the Qwen2-VL-2B \cite{wang2024qwen2}. Meanwhile, it is complementary to existing methods to boost their performance. We further verify the scalability and applicability on 7 advanced MLLMs from the series of Qwen2-VL \cite{wang2024qwen2}, InternVL2\_5 \cite{chen2024expanding} and LLaVA1\_5 \cite{liu2024visual}, where the performance improvements range from 5.45\% to 11.0\% on COCO Caption dataset \cite{chen2015microsoft}. Besides quantitative results, we conduct qualitative experiments to understand MLLM attributes about color, shape, action, location, etc. Furthermore, we illustrate that TAM is capable of explaining failure cases on the QK-VQA dataset. \cite{okvqa}, and it supports video visualization well on the STAR dataset \cite{wu2star}. In addition, it can be used as a tool to visually compare MLLMs qualitatively. Moreover, TAM shows wide applicability to explain multi-turn conversations, multi-image scenarios, and visual reasoning (see the catalog Table \ref{sup_table_catalog} in supplementary for extensive examples). Our work has three main contributions:

\begin{itemize}
    \item We introduce TAM, a novel approach to explain MLLMs, which produces high-quality activation maps for multiple generative tokens by incorporating the estimated causal inference to alleviate interference from context tokens, with the rank Gaussian filter to reduce activation noises.
    \item TAM significantly outperforms current methods for MLLM explanation, while also complementing them.
    \item TAM demonstrates wide applicability and scalability, serving as a versatile tool for object localization, model understanding, failure case analysis, video visualization, MLLM comparison, and supporting diverse scenarios (e.g., multi-turn, multi-image, visual reasoning).
\end{itemize}

\section{Related Work}
\label{sec:releated_work}

\noindent\textbf{Visual Explainability for Multimodal LLM.}
The emergence of MLLMs such as LLaVA \cite{liu2024visual}, Qwen2-VL \cite{wang2024qwen2}, InternVL \cite{chen2024expanding}, and closed-source GPT-4o \cite{hurst2024gpt} has significantly impacted various tasks involing multimodal inputs like text, images, and video. We mainly focus on the visual explanation of MLLMs via the activation maps for the input image or video frames. Regarding the modality perspective, the explainability of MLLMs is partially related to methods to explain multimodal models such as CLIP Surgery \cite{li2025closer} and transformer register \cite{darcetvision} designed for CLIP \cite{radford2021learning}; InterpreT \cite{aflalo2022vl} and Bi-Modal \cite{chefer2021generic} for BERT \cite{devlin2019bert}. From the technical aspect, conventional explanation methods are also related, including class activation map series (CAM \cite{zhou2016learning}, Grad-CAM \cite{selvaraju2017grad}, Grad-CAM++ \cite{chattopadhay2018grad}, LayerCAM \cite{jiang2021layercam}), attention-based mechanisms (LRP \cite{lapuschkin2019unmasking}, Rollout \cite{abnar2020quantifying}), the combination (AttnLRP \cite{achtibat2024attnlrp}, Grad$\times$AttnRoll \cite{chefer2021transformer}), or model-agnostic methods (LIME \cite{ribeiro2016should}, SHAP \cite{lundberg2017unified}).

Although existing methods may be valid to explain MLLMs, they are not the optimal solution. Because they are usually designed for conventional models (e.g., CNN \cite{he2016deep}, ViT \cite{dosovitskiy2020image}, CLIP \cite{radford2021learning}) that produce a single output rather than generating multiple tokens progressively in MLLMs. We have observed that these earlier context tokens interfere with later tokens by introducing redundant visual activations, as shown in Fig. \ref{fig_intro}. This phenomenon is first studied by us, solved by a novel estimated causal inference method, which is specially designed for MLLM. Notably, some workshop papers (LLaVA-CAM \cite{zhang2024redundancy} and LVLM-Interpret \cite{ben2024lvlm}) applied conventional methods \cite{omeiza2019smooth,rohekar2024causal} to explain LLaVA \cite{liu2024visual}, but they also overlooked the interference from context tokens and treated each token independently.

\noindent\textbf{Causal Model.} 
Causal inference \cite{yao2021survey} seeks to establish causal relation \cite{pearl2009causality} between variables, focusing on how changes in one variable influence another. In contrast to statistical correlation analysis, it emphasizes causal relations. Another related concept, causal intervention \cite{jiao2024causal}, involves manipulating variables using a do operator. Model-agnostic explainability methods, such as LIME \cite{ribeiro2016should} and SHAP \cite{lundberg2017unified}, can be viewed as causal interventions that explore the relationship between inputs and structured outputs by masking inputs or selecting subsets \cite{chenless}. This concept is also applied to transformers through token masking (CLEANN \cite{rohekar2024causal}) or hidden state replacement \cite{palit2023towards}. However, causal intervention for MLLMs is impractical. Because there are too many input tokens are outputs in MLLM, requiring unbearable inference times. Besides, modified input leads to a changed context every inference, which is hard to evaluate. Our method is inspired by the potential outcome model (POM) to achieve causal inference without additional model inferences. Our contribution lies in estimating the unprocessed output from multiple tokens rather than a single ready-made output, considering the uniqueness of MLLMs.

\noindent\textbf{Transformer Denoising.}
\label{sec:denoise}
Activation maps of transformers \cite{vaswani2017attention} usually present many noise activations, impeding high-quality visualization. Recent studies give their explanations about it, including attention sink \cite{xiaoefficient}, lack of register \cite{darcetvision}, and redundant features across classes \cite{li2025closer}. In MLLM, these noises still exist obversely, even system tokens already play the role of registers, or deploying the feature surgery to mitigate redundant features (See Table \ref{tab_effec}). To solve this problem practically, we aim to introduce denoising filters to mitigate it as post-processing. Since these noises belong to salt-and-pepper noise morphologically, we apply the median filter, adaptive median filter \cite{chang2008adaptive}, and the Gaussian filter. However, these filters are not the optimal solution, where the Gaussian filter keeps too much noise signal and the median filter overlooks smaller responses. Therefore, we propose the rank Gaussian filter, a novel, simple, and effective filter to denoise transformer activation. It merges ranked values within a sliding window, weighted summed by a 1-d Gaussian kernel with the center at the median rank. Besides, we improve the Gaussian kernel by the coefficient of variation for more robust results. Overall, our method addresses the activation noises in a practical aspect through a new filter.

\begin{figure*}[t]
\centering
 \includegraphics[width=1\textwidth]{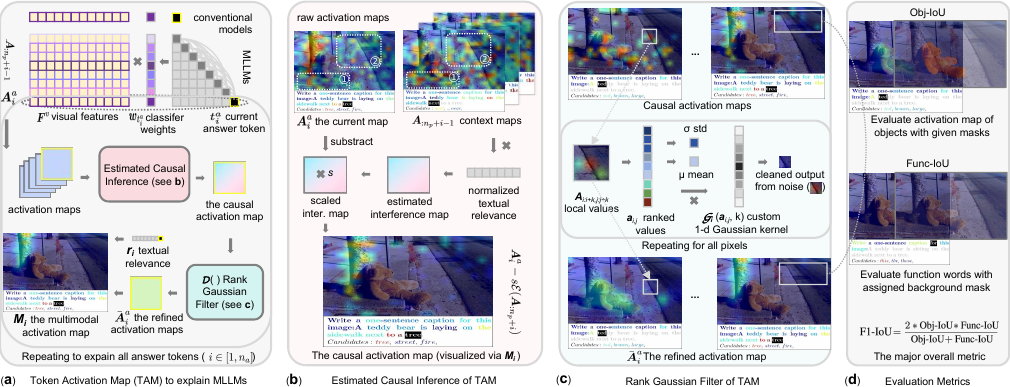}
\caption{Illustration of method. (\textbf{a}, Eq. \ref{eq_a} - Eq. \ref{eq:m_r}) The overall framework of TAM. (\textbf{b}, Eq. \ref{eq:esti} - Eq. \ref{eq:scale}) Details of the estimated casual inference module. (\textbf{d}, Eq. \ref{eq:denoise} - Eq. \ref{eq:gau}) Details of the rank Gaussian filter module. (\textbf{d}, Eq. \ref{eq:obj} - Eq. \ref{eq:f1}) Fine-grained evaluation metrics.}
\label{fig_meth}
\end{figure*}

\section{Method}
\label{sec:releated_work}
In this section, we first provide an overview of the proposed Token Activation Map (TAM), as illustrated in Fig.~\ref{fig_meth}. Next, we elaborate on two key modules within TAM: the estimated causal inference and the rank Gaussian filter. Finally, we introduce three metrics designed to enable fine-grained evaluations of the explanations for MLLMs.

\subsection{Token Activation Map}
\label{subsec:tam}
TAM aims to dipect explainable activation maps for multiple tokens generated from MLLMs by leveraging causal inference to mitigate inter-token interference. 
To be more specific, given the visual data and prompt text as the input, the MLLM is employed to progressively generate the visual features $\boldsymbol{F}^v \in \mathbb{R}^{n_v \times c}$, prompt features $\boldsymbol{F}^p \in \mathbb{R}^{n_p \times c}$ and answer features $\boldsymbol{F}^a \in \mathbb{R}^{n_a \times c}$. 
$n_v$, $n_p$, $n_a$ and $c$ indicate the number of three type of features, and the feature dimension, respectively. Next, a token classifier, instantiated as a fully connected layer, is used to generate answer tokens $\boldsymbol{t}^a =\{t^a_1\ ...\ t^a_{n_a}\}$ from these features. Prompt tokens are defined as $\boldsymbol{t}^p =\{t^p_1\ ...\ t^p_{n_p}\}$ similarly. Based on the variables described above, the visual activation maps for the prompt tokens and answer tokens are calculated as follows:
\begin{equation}
\label{eq_a}
\begin{split}
    \boldsymbol{A}^p_i &= \lfloor\boldsymbol{F}^v \boldsymbol{w}_{t^p_i}\rfloor_+, i \in [1, n_p] \\
    \boldsymbol{A}^a_i &= \lfloor\boldsymbol{F}^v \boldsymbol{w}_{t^a_i}\rfloor_+, i \in [1, n_a],
\end{split}
\end{equation}
where $\boldsymbol{A}^p_i \in \mathbb{R}^{n_v \times 1}$ and $\boldsymbol{A}^a_i \in \mathbb{R}^{n_v \times 1}$ present the activation map for the $i$-th prompt token $t^p_i$ and answer token $t^a_i$. $\boldsymbol{w}_{t^p_i} \in \mathbb{R}^{c \times 1}$ and $\boldsymbol{w}_{t^a_i} \in \mathbb{R}^{c \times 1}$ are the corresponding weight vectors within the token classifier.
$\lfloor\rfloor_+$ is the function to keep positive activations.

Then, the activation maps for prompt and answer tokens are concatenated ($\bot$) as $\boldsymbol{A}_{:n_p+n_a} = \boldsymbol{A}^p \bot \boldsymbol{A}^a$.
For the $i$-th answer token, its context token is formulated as $\boldsymbol{A}_{:n_p+i-1}$, including all earlier context maps (prompt tokens + answer tokens earlier than $i$). On top of it, the refined activation maps for answer tokens are formulated as follows:
\begin{equation}
    \bar{\boldsymbol{A}}^a_i = \lfloor\mathcal{D}(\boldsymbol{A}_i^a - s\mathcal{E}(\boldsymbol{A}_{:n_p+i-1}))\rfloor_+, i \in [1, n_a],
\label{eqn:ttl_pipeline}
\end{equation}
where $\bar{\boldsymbol{A}}^a_i$ indicates the refined activation map for the $i$-th answer token. 
The number of $i$ increases sequentially from $1$ to $n_a$.
$\boldsymbol{A}_i^a - s\mathcal{E}(\boldsymbol{A}_{:n_p+i-1})$ is the proposed estimated causal inference that produces the causal activation map with the estimation function $\mathcal{E}$ described in Sec.~\ref{subsec:eci}. $\mathcal{D}$ is the proposed rank Gaussian filter module in Sec.~\ref{subsec:rgf}. 

In the last, we concatenate ($\bot$) the refined visual activation map $\bar{\boldsymbol{A}}^a_i$ of the $i$-th answer token with the raw textual relevance $\mathbf{r}_i \in \mathbb{R}^{n_p+i-1}$ for a multimodal activation map (image or video with text) $\boldsymbol{M}_i \in \mathbb{R}^{n_v+n_p+i-1}$ as follows:
\begin{equation}
   \label{eq:m_r}
   \begin{split}
        \boldsymbol{M}_i = \mathcal{N}(\boldsymbol{\bar{A}}^a_i \bot &\mathbf{r}_i), i \in [1, n_a],
    \end{split}
\end{equation}
where $\mathcal{N}()$ is a min-max normalization function for visualizing visual and textual activations at an aligned level.

\subsection{Estimated Causal Inference}
\label{subsec:eci}
Overall, Eq.~\ref{eqn:ttl_pipeline} optimizes later visual activation maps based on visual activation maps of all earlier context tokens. To be more specific, causal inference is employed to investigate the correct causal relation between the current answer token and visual tokens, eliminating interferences from prompt tokens and preceding answer tokens. To this end, we quantify the interference map from the context tokens for the $i$-th answer token as follows:
\begin{equation}
    \label{eq:esti}
    \begin{split}
            \mathcal{E}(\boldsymbol{A}_{:n_p+i-1}) = &\sum_{k=1}^{n_p+i-1} \frac{r_i^k}{\sum \mathbf{r}_i + \epsilon} \boldsymbol{A}_k \\
            & s.t. \ r_i^k = 0, \ \text{if} \ t^a_i = t_k,
    \end{split}
\end{equation}
where $r_i^k$ is the textual relevance between the $k$-th context token $t_k$ and the current answer token $t^a_i$. The relevance is obtained from activations among textual tokens $\lfloor(\boldsymbol{F}^p \bot \boldsymbol{F}^a) \boldsymbol{w}_{t^a_i}\rfloor_+$ following Eq. \ref{eq_a}. If the current context token $t_k$ is the same as the answer token $t^a_i$, we set their relevance as $0$, \ie, $r^k_i = 0$, to avoid eliminating visual activations for tokens identical to the current answer token.

Overall, Eq.~\ref{eq:esti} aims for each answer token to encapsulate more visual information, which implies minimizing similarity to the visual semantics of previously context tokens. Meanwhile, we use the scale factor $s$ in Eq.~\ref{eqn:ttl_pipeline} to control the scale of the interference map, ensuring that the influence of the interference map is removed from the activation map at a comparable degree. Specifically, we employ the least squares method to minimize the residual between the two maps, for the optimal value of $s$, as follows:
\begin{equation}
    \label{eq:scale}
    s = \arg\min_s \sum_{j=1}^{n_v} \left( \boldsymbol{A}_{j,i}^a - s \mathcal{E}\left(\boldsymbol{A}_{j,:n_p+i-1}\right) \right)^2.
\end{equation}

After processing by the estimated causal inference module, there are still many salt-and-pepper noises in activation maps. To address this issue, we will elaborate on the solution in the following section.

\subsection{Rank Gaussian Filter}
\label{subsec:rgf}
As described in Sec. \ref{sec:denoise} the activation noise exists obviously and belongs to the salt-and-pepper type. Different from existing methods \cite{darcetvision,li2025closer} with suboptimal performances on MLLM, we address this problem in an new aspect: image denoising. Compared with existing filters like the Gaussian filter that keeps too much noise signal and the median filter which overlooks smaller responses, our rank Gaussian filter is more robust beyond existing methods. The overall equation of the rank Gaussian filter for any activation map $\boldsymbol{A}_{i,j} \in \mathbb{R}^{h \times w}$ is written as:
\begin{equation}
\label{eq:denoise}
    \begin{split}
        \mathcal{D}(\boldsymbol{A}_{i,j}, k) = \sum &\boldsymbol{A}_{i,j} \mathcal{G}(\mathbf{a}_{i,j}, k), \ \forall i \in H \ , \ \forall j \in W \\
        s.t.\ &\mathbf{a}_{i,j} = rank(\boldsymbol{A}_{i:i+k,j:j+k}).
    \end{split}
\end{equation}
Herein, $A_{i:i+k, j:j+k}$ indicates local values within a sliding window at kernel size $k$. We sort them to get a ranked array to multiply with a custom 1-d Gaussian kernel $\mathcal{G}()$. This custom kernel allows the median to have the largest weight, meanwhile aggregating signals ranked close to the median. It can be regarded as a kind of smoothed median filter to enhance the robustness weighted by the Gaussian kernel.

For the custom 1-d Gaussian kernel, we formulate it as Eq. \ref{eq:gau}. Specifically, $(i - k^2 // 2)^2$ indicates the distance from the median rank ($k^2 // 2$, size exact division 2), and $\sigma(\mathbf{a}) / \mu(\mathbf{a})$ is the coefficient of variation which is more stable than the standard division $\sigma()$ in general Gaussian kernel using the mean $\mu()$ only.
\begin{equation}
    \label{eq:gau}
    \begin{split}
        \mathcal{G}_i(\mathbf{a}, k) = e^{-\frac{(i - k^2 // 2)^2}{2(\sigma(\mathbf{a}) / \mu(\mathbf{a}))^2}}, \ \forall i \in k^2
    \end{split}
\end{equation}

\subsection{Evaluation for MLLM Explanation}
\label{subsec:metric}
Unlike conventional models with a single output, MLLMs generate multiple tokens. To evaluate how well activation maps correspond to objects, we propose two fine-grained IoU-based metrics: Obj-IoU for objects with manual masks, and Func-IoU for backgrounds. These metrics focus on plausibility instead of faithfulness (see concept differences in Supp. \ref{sec:sup_metric}). Plausibility better suits MLLMs since perturbation-based faithfulness tests \cite{achtibat2024attnlrp,chefer2021transformer} which cause inconsistent text generation, making evaluation invalid (see detailed reasons in Supp. \ref{sec:sup_metric}).

To be specific, we formulate Obj-IoU in Eq. \ref{eq:obj}, counting the IoU between activation map ($\boldsymbol{A}_i$) after binarization ($\mathcal{B}()$ using opencv OTSU auto threshold) and the ground-truth mask $\boldsymbol{G}_i$. Since the object classes may be unfixed, we count the IoU for all tokens at size $o$ over the dataset without an average on classes. We obtain the mask by matching the token to mask names. For a word or phrase containing multiple tokens, we record their max IoU for a single object. 

\begin{equation}
    \footnotesize
    \label{eq:obj}
    \text{Obj-IoU} \! = \frac{1}{o}\sum^{o}_{i=1} \frac{\mathcal{B}(\boldsymbol{A}_i) \cap \boldsymbol{G}_i}{\mathcal{B}(\boldsymbol{A}_i) \cup \boldsymbol{G}_i}
\end{equation}

Func-IoU measures the activation degree of tokens that are unreadable from the image (e.g., “is”, “and”, “so”, “the”). Higher Func-IoU indicates fewer false positives and unnecessary responses. Specifically, we record the mean binarization thresholds $b_i$ (opencv OTSU auto threshold) for all noun tokens in the same image to divide foreground and background on function words. Activation map lower than it $(\boldsymbol{A}_i < b_i)$ is the background prediction used to count IoU with the all-one matrix $\boldsymbol{J}$ (ground-truth is all background class). It is applied to total $u$ tokens over the whole dataset.
\begin{equation}
    \footnotesize
    \label{eq:func}
    \text{Func-IoU} \! = \frac{1}{u}\sum^{u}_{i=1} \frac{(\boldsymbol{A}_i < b_i) \cap \boldsymbol{J}}{(\boldsymbol{A}_i < b_i) \cup \boldsymbol{J}}
\end{equation}

We try to merge the above IoUs as the overall major metrics. We find their average is insufficient to measure biased and low-quality activation maps. For example, if one method predicts almost all tokens to be background, the Obj-IoU is very low but Func-IoU is high, forming a certain inverse ratio (e.g., 5.74\% vs. 96.5\% of Attention-Rollout \cite{abnar2020quantifying} in Table \ref{tab_sota}). One suitable solution is to use the F1 value to merge metrics with a certain inverse ratio. Thus, we have the F1-IoU to evaluate the overall plausibility stably:
\begin{equation}
    \footnotesize
    \label{eq:f1}
    \text{F1-IoU} \! = \small \frac{2 * \text{Obj-IoU} \! * \text{Func-IoU} \!}{\text{Obj-IoU} \! + \text{Func-IoU} \!}
\end{equation}

\section{Experiments}
\label{sec:releated_work}

\subsection{Setup}
\label{subsec:set}

\noindent\textbf{Datasets.} For quantitative experiments, we use datasets with texts and pixel-level annotations. Our main dataset is the COCO Caption dataset \cite{chen2015microsoft}, which draws images and masks from COCO2014 \cite{lin2014microsoft}. Since explainability methods require testing only, we use its 5K-image minival split without training. Additional datasets include GranDf \cite{rasheed2024glamm} (1K images) and OpenPSG \cite{zhou2024openpsg} (3,176-image validation set). Masks in COCO Caption and GranDf are manually annotated, while OpenPSG masks are integrated by Rasheed et al. \cite{rasheed2024glamm}. For datasets without masks, we perform qualitative visualization tests (Sec. \ref{subsec:vis}), including attribute-only images and the QK-VQA \cite{okvqa} validation set (5,046 VQA samples with textual answers). We also qualitatively evaluate the STAR video dataset \cite{wu2star} (914 videos).

\noindent\textbf{Implementations.} We use the SciPy minimize function with the BFGS method to optimize the scale factor $s$ in Eq. \ref{eq:scale}. The only hyperparameter, the kernel size $k$ in Eq. \ref{eq:gau}, is set to 3, as our observations indicate that salt-and-pepper noises typically cluster between 1 to 5 tokens. A kernel size of 3, consisting of 9 elements, is deemed suitable. Eq. \ref{eq:func} incorporates part-of-speech identification. For Func-IoU and Obj-IoU, we tag each word's part of speech using the pos\_tag function from the NLTK Python package (version 3.8.1, see tag details in Supp. \ref{sec:sup_metric}). For word matching to mask names, we employ the lemmatize function from nltk.stem. Note that the thresholds in Eq. \ref{eq:obj} and Eq. \ref{eq:func} are automatically derived from OpenCV's OTSU method, requiring no manual operation. Details of the baselines are provided in Supp. \ref{sec_supp_baselines}.

\subsection{Quantitative Results}
\label{subsec:quanti}

\vspace{-0.2cm}
\begin{table}[h]
\centering
\footnotesize
\setlength\tabcolsep{8pt}
\begin{tabular}{cc|ccc}
\hline 
ECI & RGF & Obj-IoU (\%) & Func-IoU (\%) & $\textbf{F1-IoU}$ (\%)\\
\hline
\ding{55} & \ding{55} & 21.23 & 51.93 & 30.14 \\
\ding{52} & \ding{55} & 22.41 & 69.03 & $33.84_{+3.7}$ \\
\ding{55} & \ding{52} & 24.82 & 43.34 & $31.57_{+1.33}$ \\
\ding{52} & \ding{52} & 27.37 & 68.44 & $\textbf{39.1}_{+8.96}$ \\
\hline 
\end{tabular}
\vspace{-0.2cm}
\caption{\label{tab_ablation} Ablation study on the COCO Caption \cite{chen2015microsoft} dataset using Qwen2-VL-2B \cite{wang2024qwen2}. These two modules are mutually beneficial, the combination exceeds the sum of their individual gains. ECI indicates the proposed estimated causal inference, and RGF is the rank Gaussian filter. Metrics are IoU for object words, IoU for function words, and their F1-score-like combination, respectively.}
\vspace{-0.4cm}
\end{table}

\vspace{-0.1cm}
\begin{table}[h]
\centering
\footnotesize
\setlength\tabcolsep{6pt}
\begin{tabular}{cccc}
\hline 
Setting & Obj-IoU (\%) & Func-IoU (\%) & $\textbf{F1-IoU}$ (\%) \\
\hline
Baseline & 21.23 & 51.93 & 30.14 \\
\cdashline{1-4}[1pt/5pt]
\multicolumn{4}{c}{Replace rank Gaussian filter} \\
\cdashline{1-4}[1pt/5pt]
Adaptive Median \cite{chang2008adaptive} & 25.48 & 68.2 & 37.1 \\
Median Filter & 26.01 & 68.26 & 37.67 \\
Gaussian Filter & 26.56 & 67.78 & 38.16 \\
\cdashline{1-4}[1pt/5pt]
\multicolumn{4}{c}{Replace estimated causal inference} \\
\cdashline{1-4}[1pt/5pt]
Feature surgery \cite{li2025closer} & 18.5 & 48.66 & 26.81 \\
ECI-mean & 27.84 & 49.85 & 35.72 \\
ECI-attnWeights & 27.11 & 54.61 & 36.23 \\
\cdashline{1-4}[1pt/5pt]
\textbf{Ours} & 27.37 & 68.44 & $\textbf{39.1}$ \\
\hline 
\end{tabular}
\vspace{-0.2cm}
\caption{\label{tab_effec} Effectiveness study on the COCO Caption \cite{chen2015microsoft} dataset using Qwen2-VL-2B \cite{wang2024qwen2}. ECI-mean and ECI-attnWeights are candidate implementations of ECI described in Supp. \ref{sec_supp_baselines}.}
\vspace{-0.3cm}
\end{table}

\begin{table*}[h]
\centering
\footnotesize
\setlength\tabcolsep{6pt}
\begin{tabular}{ccccccccccc}
\hline 
\multirow{2}{*}{\textbf{Method}} & \multirow{2}{*}{\textbf{Type}} & \multicolumn{3}{|c|}{\textbf{COCO Caption}} &\multicolumn{3}{c|}{\textbf{GranDf}} & \multicolumn{3}{c}{\textbf{OpenPSG}} \\
 & & \multicolumn{1}{|c}{Obj-IoU} & Func-IoU & \textbf{F1-IoU} & \multicolumn{1}{|c}{Obj-IoU} & Func-IoU & \textbf{F1-IoU} & \multicolumn{1}{|c}{Obj-IoU} & Func-IoU & \textbf{F1-IoU} \\
\hline
Grad-CAM \cite{selvaraju2017grad} & \multirow{4}{*}{Gradient} & 21.23 & 51.93 & \underline{30.14} & 17.85 & 62.15 & 27.74 & 22.93 & 48.57 & 31.15 \\
Grad-CAM++ \cite{chattopadhay2018grad} & & 19.52 & 62.83 & 29.78 & 17.3 & 73.42 & \underline{28.01} & 22.21 & 59.95 & \underline{32.41}  \\
Grad-Rollout \cite{abnar2020quantifying} & & 1.27 & 99.51 & 2.51 & 1.4 & 99.61 & 2.77 & 1.57 & 99.58 & 13.08 \\
Layer-CAM \cite{jiang2021layercam} & & 11.43 & 84.88 & 20.15 & 13.11 & 82.09 & 22.62 & 14.12 & 85.29 & 24.22  \\
\cdashline{1-11}[1pt/5pt]
Attention & \multirow{2}{*}{Attention} & 8.2 & 92.87 & 15.07 & 9.6 & 93.56 & 17.42 & 10.58 & 94.28 & 19.03 \\
Attention-Rollout \cite{abnar2020quantifying} & & 5.74 & 96.5 & 10.83 & 7.21 & 96.65 & 13.42 & 7.94 & 97.04 & 14.68 \\
\cdashline{1-11}[1pt/5pt]
CP-LRP \cite{ali2022xai} & \multirow{2}{*}{Combination} & 9.9 & 53.97 & 16.73 & 12.61 & 53.24 & 20.39 & 13.3 & 53.36 & 21.3 \\
Attn-LRP \cite{achtibat2024attnlrp} & & 9.92 & 52.41 & 16.69 & 12.15 & 52.19 & 19.72 & 12.78 & 52.26 & 20.54 \\
\cdashline{1-11}[1pt/5pt]
CAM \cite{zhou2016learning} & \multirow{3}{*}{Logit} & 21.23 & 51.93 & \underline{30.14} & 17.85 & 62.15 & 27.74 & 22.93 & 48.57 & 31.15 \\
Archi.-Surgery \cite{li2025closer} &  & 15.69 & 63.82 & 25.19 & 16.59 & 62.28 & 26.2 & 19.83 & 58.77 & 29.65 \\
\textbf{TAM (ours)} & & 27.37 & 68.44 & \textbf{39.1} & 18.65 & 88.97 & \textbf{30.83} & 26.26 & 92.99 & \textbf{40.95} \\
\hline
Archi.-Surgery \cite{li2025closer} & \multirow{4}{*}{+\textbf{ Ours}} & +4.13 & +10.68 & +6.12 & +1.41 & +26.48 & +3.74 & +1.59 & +31.92 & +5.0 \\
Grad-CAM++ \cite{chattopadhay2018grad} & & +2.83 & +3.91 & +3.7 & +1.26 & +5.39 & +2.04 & +2.75 & +9.44 & +4.3 \\
Layer-CAM \cite{jiang2021layercam} & & +5.04 & -7.95 & +6.98 & +2.79 & +0.52 & +\textbf{4.05} & +6.62 & -2.49 & +8.95 \\
CAM \cite{zhou2016learning} & & +6.14 & +16.52 & +\textbf{8.96} & +0.79 & +26.82 & +3.09 & +3.33 & +44.42 & +\textbf{9.8} \\
\hline 
\end{tabular}
\caption{\label{tab_sota} Comparison with SoTA methods using Qwen2-VL-2B on diverse datasets. We adopt the “Logit” type that uses the classifier weights in Eq. \ref{eq_a}, without back-propagation in “Gradient” and “Combination” (Gradient + Attention). Besides, we support FlashAttention \cite{dao2022flashattention} and SdpaAttention, which do not return attention weights, while “Attention” and “Combination” rely on it. Our TAM is also complementary to existing methods with gains marked by “+”. Note, CAM and Grad-CAM are equivalent as discussed in Supp. \ref{sec_supp_baselines}. The major metric is the F1-IoU (\%) to reflect the overall result, merged from IoUs for Object words (Obj-IoU) and function words (Func-IoU). }
\end{table*}

\begin{table*}[h]
\centering
\footnotesize
\setlength\tabcolsep{9pt}
\begin{tabular}{cccccccccc}
\hline 
\multirow{2}{*}{\textbf{CAM \cite{zhou2016learning} +Ours}} & \multicolumn{3}{|c|}{\textbf{COCO Caption}} &\multicolumn{3}{c|}{\textbf{GranDf}} & \multicolumn{3}{c}{\textbf{OpenPSG}} \\
& \multicolumn{1}{|c}{Obj-IoU} & Func-IoU & \textbf{F1-IoU} & \multicolumn{1}{|c}{Obj-IoU} & Func-IoU & \textbf{F1-IoU} & \multicolumn{1}{|c}{Obj-IoU} & Func-IoU & \textbf{F1-IoU} \\
\hline
LLaVA1\_5-7B \cite{liu2024visual} & +4.47 & +18.27 & +7.97 & +0.65 & +11.67 & +2.47 & +3.47 & +9.51 & +5.16 \\
LLaVA1\_5-13B \cite{liu2024visual} & +4.3 & +7.31 & +5.45 & +0.76 & +7.03 & +2.11 & \textbf{+4.22} & +11.51 & +6.37 \\
InternVL2\_5-2B \cite{chen2024internvl} & +5.44 & +19.48 & +8.57 & +2.2 & \textbf{+48.29} & \textbf{+8.47} & +3.24 & +40.45 & +8.65 \\
InternVL2\_5-4B \cite{chen2024internvl} & +3.53 & +22.17 & +7.14 & +1.62 & +45.19 & +7.56 & +2.2 & \textbf{+55.05} & +10.85 \\
InternVL2\_5-8B \cite{chen2024internvl} & +5.38 & +2.13 & +6.93 & \textbf{+3.53} & +28.53 & +7.02 & +3.27 & +26.54 & +6.44 \\
Qwen2-VL-2B \cite{wang2024qwen2} & \textbf{+6.14} & +16.52 & +8.96 & +0.79 & +26.82 & +3.09 & +3.33 & +44.49 & +9.8 \\
Qwen2-VL-7B \cite{wang2024qwen2} & +5.62 & \textbf{+29.4} & \textbf{+11.01} & +1.29 & +22.54 & +3.4 & +3.53 & +46.94 & \textbf{+11.15} \\
\cdashline{1-10}[1pt/5pt]
Mean Improvements & +4.98 & +16.47 & +8.01 & +1.55 & +27.15 & +4.88 & +3.32 & +33.5 & +8.35 \\
\hline 
\end{tabular}
\caption{\label{tab_mllm} Explainability improvements of TAM on diverse MLLMs and datasets. Specific results are listed in Table \ref{tab_sup_mllm} of Supp. \ref{sec_supp_mllms}.}
\end{table*}

\noindent\textbf{Ablation and Effectiveness.} We present the ablation study results in Table \ref{tab_ablation}, which indicates that each module of the TAM is valuable. Specifically, the estimated causal inference significantly improves Func-IoU by reducing context interference on function words, while the rank Gaussian filter performs better for object words. Importantly, these modules are mutually beneficial; their combination leads to an F1-IoU improvement of 8.96\%, exceeding the total of their individual contributions (3.7\% and 1.33\%).

We further assess the effectiveness of these modules by comparing them with other methods, as shown in Table \ref{tab_effec}. Replacing our proposed rank Gaussian filter with alternatives yields lower performance, such as the adaptive median filter \cite{chang2008adaptive}, the Gaussian filter, and the median filter. Although the differences are less pronounced than those seen with the causal inference module, denoising remains a challenging area. Our method outperforms feature surgery \cite{li2025closer} (which operates on the class dimension) by 12.29\% and also surpasses other ECI implementations in Supp. \ref{sec_supp_baselines}.

\noindent\textbf{Comparison with SoTA Methods.} Table \ref{tab_sota} presents a comparison between TAM and SoTA methods. Our TAM outperforms all others in the major F1-IoU and Obj-IoU metrics. While some methods, such as Grad-Rollout \cite{abnar2020quantifying} and Attention, achieve high Func-IoU by producing a limited number of activations, their performance on other metrics remains low. The second-best method is Grad-CAM++ \cite{chattopadhay2018grad} on the COCO Caption \cite{chen2015microsoft} and OpenPSG \cite{zhou2024openpsg} datasets, while CAM \cite{zhou2016learning} and Grad-CAM \cite{selvaraju2017grad} rank second on GranDf \cite{rasheed2024glamm}. Our method surpasses these approaches by 8.96\%, 2.82\%, and 8.54\% on these datasets, respectively. Furthermore, TAM is complementary to existing methods and enhances them significantly, with maximum gains of 6.62\%, 44.42\%, and 9.8\% across the three metrics, respectively.

\begin{figure*}[t]
\centering
 \includegraphics[width=1\textwidth]{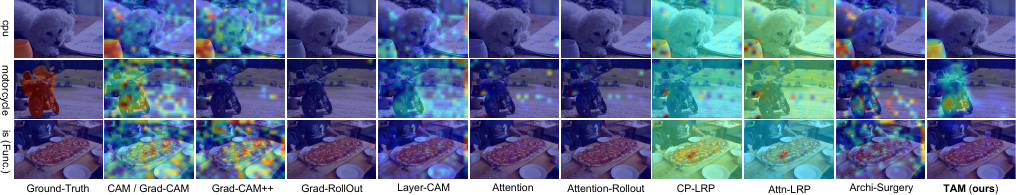}
\caption{Visual comparison between TAM and SoTA methods on the COCO Caption dataset \cite{chen2015microsoft} using the Qwen2-VL-2B \cite{wang2024qwen2} model. The “is” in the last row is a functional word with background ground-truth. More examples are shown in Supp. \ref{sec:sup_vis_methods} including complex cases.}
\label{fig_vis_methods}
\end{figure*}

\noindent\textbf{Applicability and Scalability.} We evaluate the applicability on three prominent MLLMs, as shown in Table \ref{tab_mllm}, and assess its scalability across various model sizes. The results in Table \ref{tab_sota} indicate that TAM consistently outperforms the baseline across all MLLMs and model sizes by significant margins. This demonstrates that TAM can be effectively applied to a range of MLLMs while providing superior explainability compared to existing methods. Moreover, TAM supports MLLMs comparison, complementing quantitative results. It introduces a new evaluation perspective by highlighting visual alignment quality, as illustrated in Fig. \ref{fig_vis_mllms}.

\subsection{Qualitative Results}
\label{subsec:vis}

\noindent\textbf{Comparison of Explainability Methods.}

We perform comparisons between the proposed TAM and SoTA explanation methods, as illustrated in Fig. \ref{fig_vis_methods}. The visualization results clearly show that TAM more accurately highlights target objects and renders fewer false activations on function words. In contrast, attention-based methods \cite{abnar2020quantifying,ali2022xai,achtibat2024attnlrp} produce only a limited number of activations, while gradient-based methods \cite{selvaraju2017grad,chattopadhay2018grad} present too many correlated responses and noise, performing significantly worse than our approach. Additionally, we present more complex examples in Supp. \ref{sec:sup_vis_methods}, where TAM notably outperforms existing SoTA methods. TAM excels at explaining multiple tokens of MLLMs. In Supp. \ref{sec:sup_vis_all_tokens}, results show how TAM focuses on important tokens, whereas the baseline methods generate many redundant activations.

\begin{figure}[t]
\centering
 \includegraphics{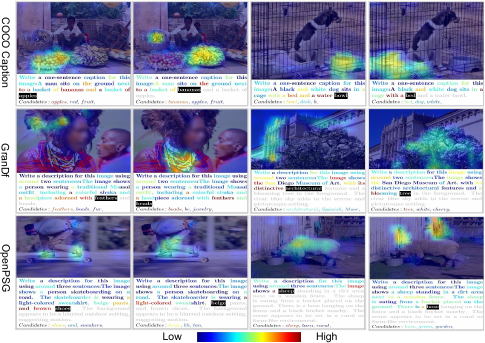}
\caption{TAM presents high-quality localization results on diverse datasets \cite{chen2015microsoft,rasheed2024glamm,zhou2024openpsg} under the Qwen2-VL-7B \cite{wang2024qwen2} model.}
\label{fig_vis_datasets}
\end{figure}

\noindent\textbf{High-quality Localization Results.}
We visualized the results using the multimodal activation map in Eq. \ref{eq:m_r} (see detailed examples in Supp. \ref{sec:sup_example}). The results demonstrate that TAM effectively locates objects across diverse datasets, as shown in Fig. \ref{fig_vis_datasets}. These findings indicate that TAM enhances MLLMs with localization capabilities, which could be potentially beneficial for various downstream tasks, including segmentation \cite{li2025closer}, object counting \cite{Shi_2024_WACV}, anomaly detection \cite{chen2024clip}, image editing \cite{bar2022text2live}, autonomous driving \cite{10656599}, and applications in the medical field \cite{wang2025interpretable,ding2024hia}, which require pixel-level activations.

\begin{figure}[t]
\centering
 \includegraphics[width=0.47\textwidth,height=0.34\textwidth]{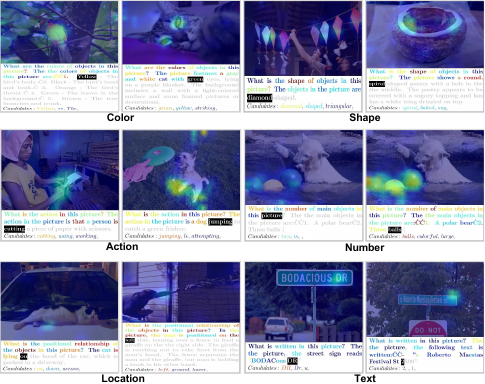}
\caption{TAM supports to explain and analyze diverse attributes of MLLM (Qwen2-VL-7B). See extensive examples in Supp. \ref{sec:sup_attributes}.}
\label{fig_vis_attri}
\end{figure}

\begin{figure*}[t]
\centering
 \includegraphics[width=1\textwidth]{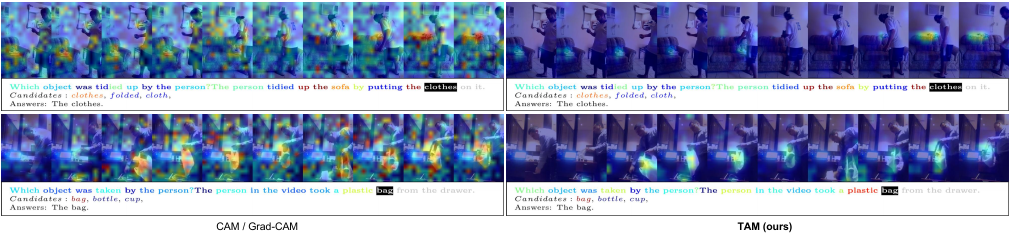}
\caption{TAM significantly improves the quality of video visualization, with much fewer redundant activation and noises on STAR \cite{wu2star} dataset using Qwen2-VL-2B \cite{wang2024qwen2}. Please see extensive examples in Supp. \ref{fig_sup_vis_videos} and failure cases analysis for videos in Fig. \ref{fig_sup_vis_video_failure_case}.}
\label{fig_vis_videos}
\end{figure*}

\noindent\textbf{Visualization for Various Attributes.}
We apply TAM to explain various attributes of MLLMs, specifically focusing on fine-grained token types as Fig. \ref{fig_vis_attri}. The results indicate that TAM effectively supports the explanation of attributes such as color, shape, action, number, location, and text. Notably, the representation of numbers is less pronounced compared to object localization, suggesting that some attributes may be encoded within related object tokens. In contrast, other attributes are more prominently featured.

\begin{figure}[t]
\centering
 \includegraphics{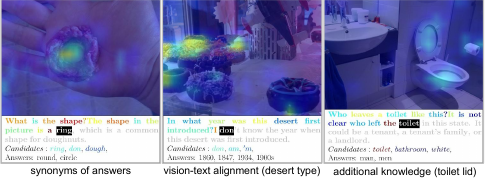}
\caption{TAM supports failure cases analysis on QK-VQA \cite{okvqa}. See more cases in Supp. about \textbf{VQA} (Fig. \ref{fig_sup_vis_failure_case}), \textbf{videos} (Fig. \ref{fig_sup_vis_video_failure_case}), visual \textbf{reasoning} (Fig. \ref{fig_sup_vis_reasoning}), and \textbf{multi-turn} conversation (Fig. \ref{fig_sup_vis_multi_turn2}).}
\label{fig_vis_cases}
\end{figure}

\noindent\textbf{Explaining Failure Cases.}
We apply TAM to analyze failure cases in the QK-VQA dataset \cite{okvqa} as Fig. \ref{fig_vis_cases}. Our findings indicate that while the model successfully locates objects, it struggles to align them with specific knowledge. For instance, the model can identify the desert and focus on the key prompt (``year'') when outputting ``don't know.'' It suggests a misalignment between the desert and the year, rather than recognition or prompt issues. Besides, some responses consist of synonyms, hypernyms, or hyponyms of the correct answers, resulting in mismatches. For more case studies, please refer to the catalog in Supp. Table \ref{sup_table_catalog}.

\noindent\textbf{Video Visualization.}
The proposed TAM excels at video visualization, where there are much fewer redundant activations and noises as shown in Fig. \ref{fig_vis_videos}. Our method provides very clear results, while the conventional CAM makes it hard to see the video content without obvious highlights.

\begin{figure}[t]
\centering
 \includegraphics{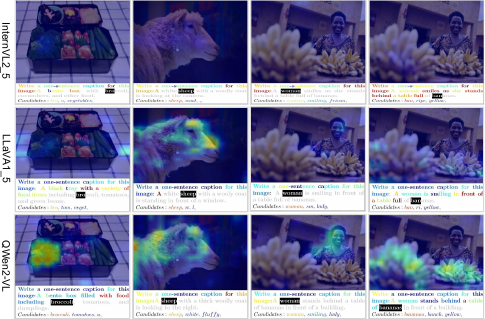}
\caption{TAM enables qualitative comparison among MLLMs. Extensive examples are given in Fig. \ref{fig_sup_vis_mllms_attri} and Fig. \ref{fig_sup_vis_mllms_coco} in Supp. \ref{sec_supp_mllms}.}
\label{fig_vis_mllms}
\end{figure}

\noindent\textbf{A Tool to Visually Comparison MLLMs.}
TAM can be used as a tool to compare MLLMs qualitatively. If an MLLM exhibits superior activation maps, it can be inferred that its explainability and degree of vision alignment are also better, offering valuable pixel-level predictions at no additional cost. As shown in Fig. \ref{fig_vis_mllms}, Qwen2-VL-7B \cite{wang2024qwen2} shows greater explainability than LLaVA1\_5-7B \cite{liu2024visual} and InternVL2\_5-8B \cite{chen2024internvl}. Furthermore, we observe that InternVL tends to focus more on text, resulting in weaker activation levels; in some cases, this performance is even better than that of LLaVA (e.g., with the broccoli). More extensive examples can be found in Supp. \ref{sec_supp_mllms}.

\section{Conclusion}
This work focuses on the unique nature of MLLMs in visual explainability, which generates multiple tokens progressively, complicating the explanation process. In response, we introduced the Token Activation Map (TAM), a novel method that utilizes estimated causal inference to mitigate interferences from context tokens, with a rank Gaussian filter to reduce activation noise, thereby providing clearer visual explanations. Our results demonstrate that TAM significantly outperforms existing SoTA methods, offering high-quality visualizations applicable to diverse scenarios.

Despite the success, we focus on visual inputs, other modalities, such as audio, remain under-explored. Additionally, interpreting model decisions is an extensible further aspect. The potential applications of our TAM method are extensive, including open-vocabulary segmentation, detection, image grounding, anomaly detection, remote sensing, and medical fields that require pixel-level activations. Our work lays a foundation for advancing their explainability and practical utility across various domains.

\section*{Acknowledgment}
This work was supported by a research grant from the Joint Research Scheme (JRS) under the National Natural Science Foundation of China (NSFC) and the Research Grants Council (RGC) of Hong Kong (Project No. N\_HKUST654/24), as well as a grant from the RGC of the Hong Kong Special Administrative Region, China (Project No. R6005-24).

{
    \small
    \bibliographystyle{ieeenat_fullname}
    \bibliography{main}
}

\clearpage
\maketitlesupplementary
\appendix

\captionsetup{justification=raggedright, singlelinecheck=false}
\begin{table}
\vspace{2cm}
\renewcommand{\arraystretch}{2}
\footnotesize
\begin{tabular}{|l|l|}
\hline 
\textbf{Section and Textual Content} & \textbf{Graphic and Tabular Content} \\
\hline
Supp. \ref{sec_sup_catalog}: Tabular Catalog of the Supplementary. & Table \ref{sup_table_catalog}: The \textbf{catalog} for quick reference. \\
\hline
Supp. \ref{sec:sup_example}: Examples of Multimodal Activation Map & Fig. \ref{fig_sup_vis_example}: \textbf{How to read} these visualization examples? \\
\hline 
Supp. \ref{sec:sup_vis_all_tokens}: TAM to Explain All Generated Tokens. & Fig. \ref{fig_sup_vis_all_tokens}: A \textbf{complete example} explaining all tokens.\\
\hline 
Supp. \ref{sec_supp_baselines}: Implementation of Baselines. & N/A, MLLM and explainability \textbf{baselines}. \\
\hline
Supp. \ref{sec:sup_metric}: Details of Metrics. & N/A, Invalid faithfulness evaluation for MLLM and other metric details. \\
\hline
Supp. \ref{sec:sup_stas}: Analysis About Motivation. & N/A, Statistical test and causal validation. \\
\hline 
Supp. \ref{sec:sup_vis_methods}: Extensive Cases About Method Comparison. & Fig. \ref{fig_sup_vis_method1} \& Fig. \ref{fig_sup_vis_method2}: TAM exceeds existing \textbf{SoTA methods} in extensive visualizations. \\
\hline 
Supp. \ref{sec:sup_vis_eci}: Visual Comparison About Causal Inference  & Fig. \ref{fig_sup_vis_eci}: How does the \textbf{estimated causal inference} work in visualization? \\
\hline 
Supp. \ref{sec:sup_vis_denoise}: Visual Comparison Among Denoise Filters & Fig. \ref{fig_sup_vis_denoise}: How does the \textbf{rank Gaussian filter} work in visualization? \\
\hline 
Supp. \ref{sec:sup_vis_ablation}: Visualization of Ablation Study. & Fig. \ref{fig_sup_vis_ablation}: Two involved modules are \textbf{mutually beneficial}. \\
\hline 
Supp. \ref{sec_supp_mllms}: Explainability Results on Diverse MLLMs. & Table \ref{tab_sup_mllm}: MLLM \textbf{quantitative results}; Fig. \ref{fig_sup_size_effect} relation between \textbf{model size} and explainability. \\
\hline 
Supp. \ref{sec:sup_vis_mllms}: TAM for MLLM Visual Comparison. & Fig. \ref{fig_sup_vis_mllms_attri} \& Fig. \ref{fig_sup_vis_mllms_coco}: TAM supports visual \textbf{comparison among MLLMs} about attributes. \\
\hline 
Supp. \ref{sec:sup_attributes}: Extensive Cases About Attributes analysis. & Fig. \ref{fig_sup_vis_attri1} \& Fig. \ref{fig_sup_vis_attri2} \& Fig. \ref{fig_sup_vis_attri3} \& Fig. \ref{fig_sup_vis_attri_comp}: Explaining \textbf{fine-grained attributes} beyond SoTA. \\
\hline 
Supp. \ref{sec:sup_bias}: TAM for Biased Scenario. & Fig. \ref{fig_sup_vis_biased}: TAM supports \textbf{biased error analysis}. \\
\hline 
Supp. \ref{sec:sup_failure}: Extensive Failure Cases Study. & Fig. \ref{fig_sup_vis_failure_case} \& Fig. \ref{fig_sup_vis_video_failure_case}: TAM supports \textbf{failure cases analysis} for images and videos. \\
\hline 
Supp. \ref{suc:sup_vis_vqa}: Extensive Success VQA Examples. & Fig. \ref{fig_sup_vis_vqa}: Explanation result on the \textbf{VQA} dataset. \\
\hline 
Supp. \ref{sec:sup_vis_video}: Examples About Video Visualization. & Fig. \ref{fig_sup_vis_videos}: Clearer \textbf{video visualizations} with fewer redundant activations and noises. \\
\hline 
Supp. \ref{sup_sec_reasoning}: Corner Case About Reasoning. & Fig. \ref{fig_sup_vis_reasoning}: TAM supports failure case analysis for \textbf{visual reasoning}. \\
\hline 
Supp. \ref{sec:sup_multi_img}: TAM for Multi-image Conversation. & Fig. \ref{fig_sup_vis_multi_img}: High applicability on \textbf{multi-image} conversation. \\
\hline 
Supp. \ref{sec:sup_multi_turn}: TAM for Multi-turn Conversation. & Fig. \ref{fig_sup_vis_multi_turn} \& Fig. \ref{fig_sup_vis_multi_turn2}: TAM supports \textbf{multi-turn conversation} about attributes and case study. \\
\hline 
\end{tabular}
\caption{Tabular catalog of the supplementary.}
\label{sup_table_catalog}
\end{table}
\clearpage

\section{Tabular Catalog of the Supplementary}
\label{sec_sup_catalog}
In this supplementary material, we primarily provide extensive qualitative results to demonstrate the effectiveness and wide applicability of TAM. These sections include comparisons with state-of-the-art (SoTA) methods, visualizations about ablation study, attribute explanation, failure case analysis, VQA examples, video visualizations, MLLM comparisons, reasoning analysis, multi-turn conversation, multi-image input, as well as some quantitative results and baseline descriptions. To enhance readability given the extensive content, we provide a tabular catalog in Table \ref{sup_table_catalog} for quick reference.

\section{Examples of Multimodal Activation Map}
\label{sec:sup_example}

In this section, we present a high-resolution example accompanied by detailed captions to facilitate the explanation of the multimodal activation map defined in Eq. \ref{eq:m_r}. The primary element is the activation map at the top, which reflects the degree of vision-text alignment and serves to visually explain the MLLM. All multimodal activation maps in this paper adhere to a consistent format, and we provide high-quality images; please zoom in if any example appears too small to read.

The visual activations and textual relevances are normalized to the same scale as specified in Eq. \ref{eq:m_r}, allowing for a direct comparison between the two modalities to identify where the model focuses—whether on the image or the context. The text is colored by tokens, with some words represented by multiple tokens marked in different colors. The answers following the target are not visible for the current explained token, and are therefore colored in gray. The colors of the candidate responses reflect the prediction confidence of the top three tokens corresponding to the target, which can be useful for analyzing failure cases through potential predictions and confidence levels associated with each token.

\begin{figure}[ht]
\centering
 \includegraphics{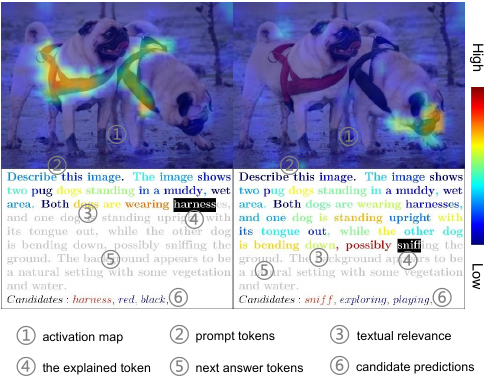}
\caption{A high-resolution example of the multimodal activation map. This image is processed by the Qwen2-VL-2B model \cite{wang2024qwen2}. There are five components to visualize, including the visual activation map, prompt tokens, textual relevance, the explained target token, next answer tokens, and its top predictions (top 3). The colors indicate the corresponding response degree.}
\label{fig_sup_vis_example}
\end{figure}

\begin{figure*}[h]
\centering
 \includegraphics[width=1\textwidth]{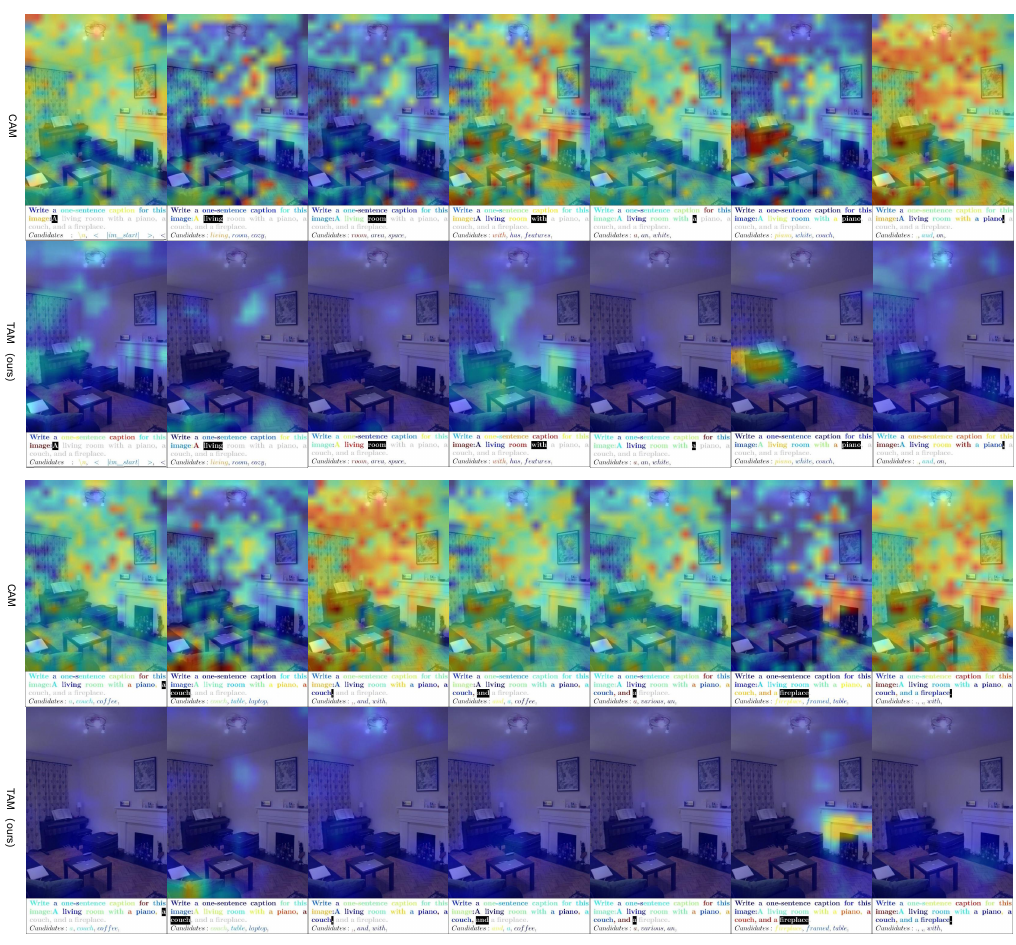}
\caption{Visualization of one example for all generated tokens on the Qwen2-VL-2B \cite{wang2024qwen2} model. \emph{The proposed TAM shows more accurate object localization ability beyond conventional CAM, with much fewer redundant activations in an overall view}.}
\label{fig_sup_vis_all_tokens}
\end{figure*}

\section{TAM to Explain All Generated Tokens}
\label{sec:sup_vis_all_tokens}
The proposed TAM demonstrates a significant advantage in explaining multiple generated tokens from MLLMs, in contrast to conventional models that typically focus on a single output. We depict all multimodal activation maps in Fig. \ref{fig_sup_vis_all_tokens} and support it. The figure clearly shows that TAM produces considerably fewer redundant activations, particularly for non-object words, thanks to the proposed estimated causal inference.

For instance, the activations for the function word “with” and the punctuation mark “.” in the Class Activation Map (CAM) are extremely high, overshadowing object activations. Additionally, these visual activations often exceed those of text tokens, resulting in excessive redundant activations. In comparison, our TAM is much clearer and focuses more on important objects. These results suggest that TAM produces closer explanations to the understanding of humans than CAM, where words from the image are highlighted while those related to texts show much fewer responses. Besides, there are much fewer activations showing higher visualization quality as well.

\section{Implementation of Baselines} 
\label{sec_supp_baselines}
We conduct experiments on various MLLMs, including Qwen2-VL-2B \cite{wang2024qwen2}, Qwen2-VL-7B, LLaVA1\_5-7B \cite{liu2024visual}, LLaVA1\_5-13B, InternVL2\_5-2B \cite{chen2024expanding}, InternVL2\_5-4B, and InternVL2\_5-8B implemented by transformers using weights from huggingface. For Qwen2-VL all the models the weights are the Instruct version (e.g., Qwen2-VL-2B from huggingface ``Qwen/Qwen2-VL-2B-Instruct''). The example of LLaVA1\_5 weights is from ``llava-hf/llava-1.5-7b-hf'' and ``thisisiron/InternVL2\_5-2B'' for InterVL2\_5. Due to device limitations, very large MLLMs are not used. For image resolution, Qwen2-VL supports raw image size, while LLaVA1\_5 and InternVL2\_5 fix image sides at 336 and 448, respectively. For the implementation of video caption on Qwen2-VL \cite{wang2024qwen2}, we extract 10 frames from a short video and repeat frames for the number of temporal\_patch\_size to ensure each frame has its own activation, instead of activations from other frames. We use the same prompts for the involved MLLMs. These prompts are set according to the average length of captions. For COCO Caption \cite{chen2015microsoft} the prompt is “Write a one-sentence caption for this image:”, and the prompts for GranDf \cite{rasheed2024glamm} and OpenPSG \cite{zhou2024openpsg} are “Write a description for this image using around two sentences:”, “Write a description for this image using around three sentences:”, respectively. For attribute analysis, the prompt is set like “What is the [attribute] in this picture?”. In addition, we use the prompts provided from the QK-VQA \cite{okvqa} and STAR \cite{wu2star} datasets, which vary according to the images.
 
 For the explanation baselines \cite{selvaraju2017grad,achtibat2024attnlrp,chattopadhay2018grad,jiang2021layercam,li2025closer}, we implement them referring to their official codebases. To obtain attention weights for attention-based explainability methods (such as attention weights and Rollout \cite{abnar2020quantifying}), we rewrite the SdpaAttention in PyTorch. This is necessary because the original implementations of SdpaAttention and FlashAttention do not provide attention weights. For the CP-LRP \cite{ali2022xai} and AttnLRP \cite{achtibat2024attnlrp}, modules of MLLM are replaced by the official implementations of AttnLRP to back-propagate the relevance from output to visual tokens. Besides, these methods need to close the kv\_cache to maintain gradients for key and value. Note, the Grad-CAM \cite{selvaraju2017grad} is equivalent to CAM. Because the weights for the activation map are derived from the classifier weight in MLLMs. Since there is only a feature vector without the pooling and other structures. The gradient is fully dependent on the classifier weight at the same ratio among channels. Since they are the same, we apply the classifier weight of CAM to achive TAM to avoid extra back-propagation in Grad-CAM. Other implementation methods of estimated causal inference include the mean of context maps (mean), using the attention weights as relevance in Eq. \ref{eq:esti} (AttnWeights).

\section{Details of Metrics.}
\label{sec:sup_metric}

The metrics are based on the part of speech using the pos\_tag function from the NLTK Python package. The specific tags are  “NN”, “NNS”, “NNP”, and “NNPS” for Obj-IoU. Function words are identified by the tags: “CC”, “DT”, “EX”, “MD”, “POS”, “PRP”, “PRP\$”, “UH”, “WDT”, “WP”, “WP\$”, and “WRB”. Notably, we exclude the tags “IN” and “CD” from function words, as they pertain to location and quantity.

Besides the IoU based metric, in Sec. \ref{subsec:metric}, we have discussed the difference between the used plausibility test (how accurately it reflects the true reasoning process), compared with another widely used faithfulness (how accurately it reflects the true reasoning process). While the faithfulness is not suitable for MLLM evaluation. Because the perturbation tests \cite{achtibat2024attnlrp,chefer2021transformer} of the faithfulness metric alter the generated texts every time, resulting in inconsistent generated texts that are not stable to evaluate raw generated texts of MLLM. Specifically, masking different input regions in the faithfulness test drastically changes MLLM output tokens, invalidating observations of "decision-making" tied to a fixed class. In conventional models, input changes affect a fixed class's confidence, but in MLLMs, it causes vanished tokens or shifted context, making confidence comparisons invalid. Besides, its cost is unacceptable, which needs N times repeated inferences (N = token number $\times$ regions ratios). 

Another consideration regarding the metrics is the variation in response levels. Specifically, our ECI involves a subtraction operation between activation maps, which can lead to a lower overall intensity compared to the original responses. We did not overlook this limitation when design the metrics; instead, we implemented a straightforward operation to penalize excessive discrepancies. Specifically, we use the response map of the first prompt token in place of the first generated token in evaluation. Since the first prompt token does not have any earlier text tokens, the map does not incorporate the ECI and reflects the original response level. If the response level of the altered map significantly differs from the processed maps, it can result in inappropriate background thresholds, thereby diminishing the Func-IoU metric. For instance, if the background threshold processed after ECI is 0.1, it may be too low for this map, leading to false positives and consequently affecting the metrics. Detailed operations can be referenced in our open-source code.

\section{Analysis About Motivation}
\label{sec:sup_stas}

In Fig. \ref{fig_intro}(c), we randomly pair CAMs and count their L1 distance against text correlation. Higher text correlation corresponds to lower distance, indicating concurrent interferences. In this section, we provide a statistical test to support it. Specifically, the added statistical test is the Pearson correlation at -0.16 with p-value of 1.5E-30. Since most pairs are not related in the random pairing, the correlation is not strong. When pairing the most related tokens, the Pearson correlation comes to -0.359 (p-value 7.9E-32). It confirms that the negative correlation is evident.

We also conducted a causal validation for the causal inference. In this paper, our ECI is based on the potential outcome model (POM). The used causal validation for this model is the Placebo test. Specifically, we validate it by replacing the target CAM to a random earlier CAM as the placebo (not the observed target), and then record the results drop. The Obj-IoU reduced to 6.2\% on COCO Caption and 4.4 times lower than the raw result, suggesting the causal effect is significant.

\section{Extensive Cases about Method Comparison}
\label{sec:sup_vis_methods}
\begin{figure*}[h]
\centering
 \includegraphics[width=1\textwidth]{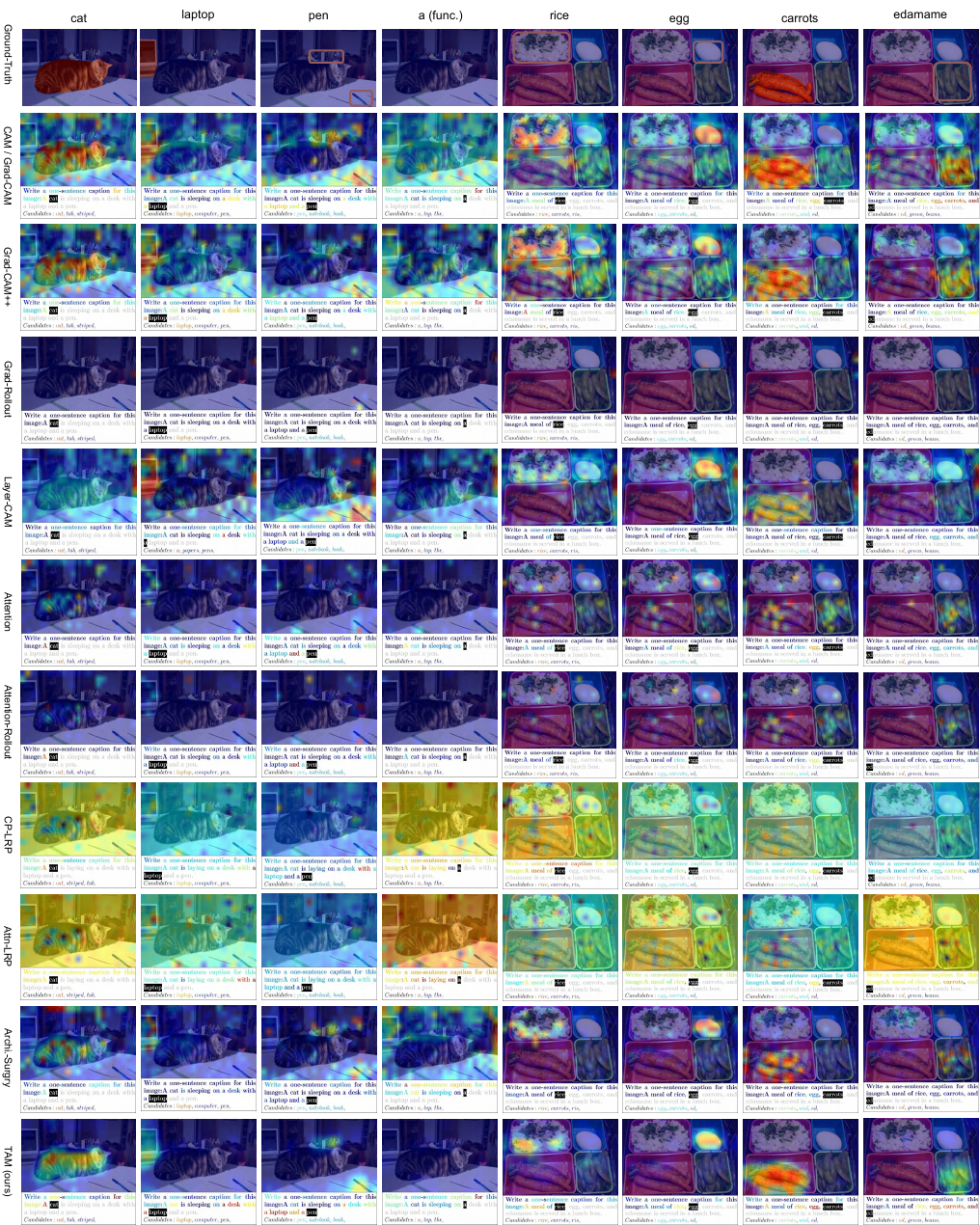}
\caption{Visual comparison between our TAM and SoTA methods on the COCO Caption dataset \cite{chen2015microsoft} using the Qwen2-VL-2B \cite{wang2024qwen2} model. Objects without ground-truth are marked by red boxes. \textbf{TAM performs best beyond previous SoTA methods}. "CAM / Grad-CAM" indicates CAM \cite{zhou2016learning} and Grad-CAM \cite{selvaraju2017grad} are equivalent for MLLM as discussed in Supp. \ref{sec_supp_baselines}. }
\label{fig_sup_vis_method1}
\end{figure*}

\begin{figure*}[h]
\centering
 \includegraphics[width=1\textwidth]{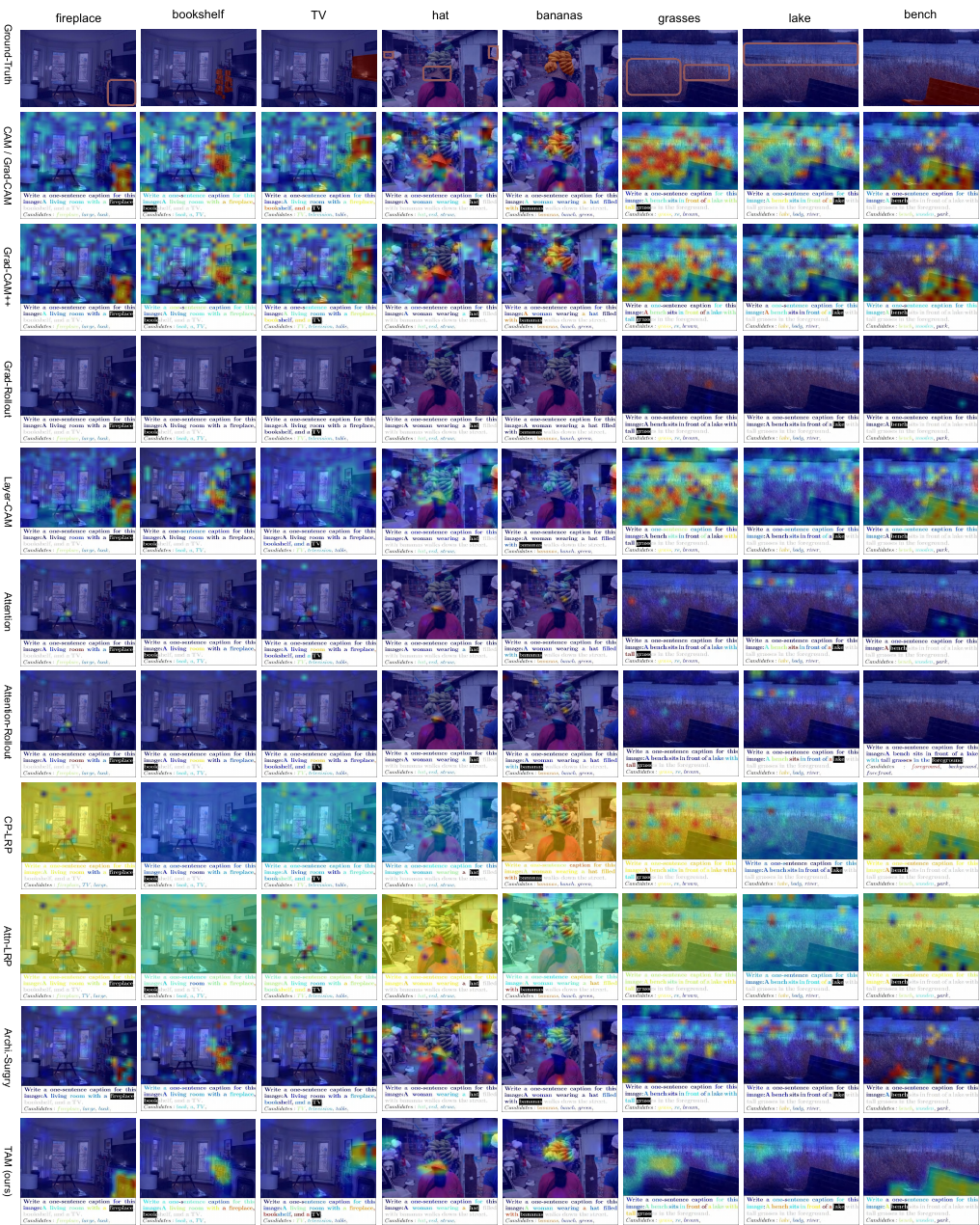}
\caption{Visual comparison between our TAM and SoTA methods on the COCO Caption dataset \cite{chen2015microsoft} using the Qwen2-VL-2B \cite{wang2024qwen2} model. Objects without ground-truth are marked by red boxes. “func.” indicates function words assigned as the background class. \textbf{TAM performs best beyond previous SoTA methods}.}
\label{fig_sup_vis_method2}
\end{figure*}

In addition to the visual comparison presented in Fig. \ref{fig_meth}, we offer more complex examples in Fig. \ref{fig_sup_vis_method1} and Fig. \ref{fig_sup_vis_method2} within this section. The findings are consistent with those discussed in Sec. \ref{subsec:vis}: the proposed TAM significantly outperforms existing explainability methods.

Specifically, TAM generates fewer redundant activations and exhibits less noise compared to gradient-based methods \cite{selvaraju2017grad,chattopadhay2018grad}. Moreover, it effectively locates objects, contrasting with the scattered activations seen in attention-based methods (e.g., Attention, Attention-Rollout \cite{abnar2020quantifying}, CP-LRP \cite{ali2022xai}, AttnLRP \cite{achtibat2024attnlrp}). These results indicate that TAM enhances the localization capabilities of MLLMs, even in complex scenarios. Consequently, TAM can be integrated into existing MLLMs without requiring grounding abilities, thereby facilitating a wide range of potential downstream tasks without additional supervision or alignment.

\section{Visual Comparison About Causal Inference}
\label{sec:sup_vis_eci}
\begin{figure*}[h]
\centering
 \includegraphics[width=1\textwidth]{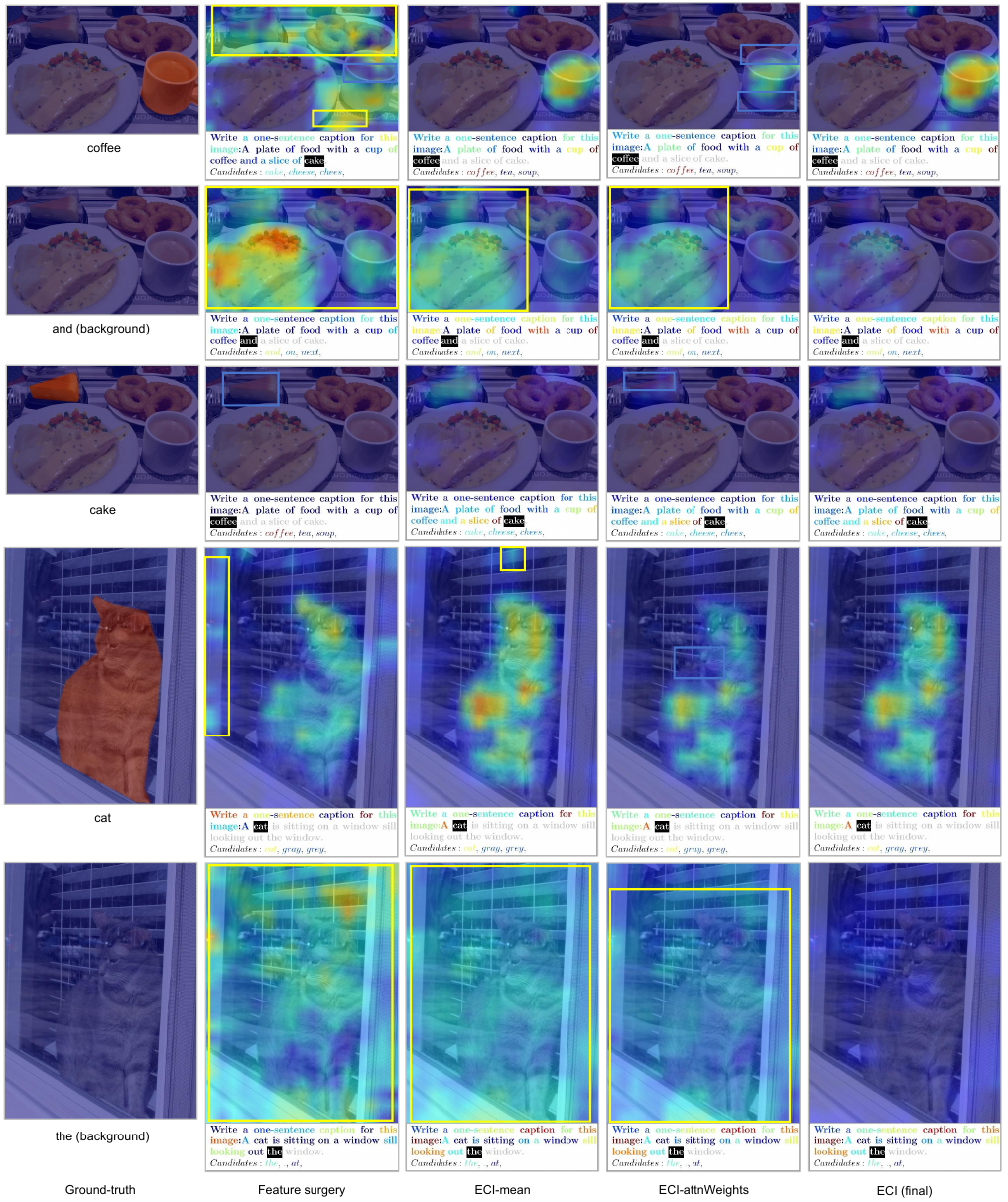}
\caption{\textbf{The proposed estimated causal inference is well-designed beyond other methods and implementations}. The yellow boxes indicate correlated activations, and the blue boxes mean missed activations. Feature surgery \cite{li2025closer} is designed for CLIP \cite{radford2021learning} to mitigate redundant features along the class dimension, while ECI-mean and ECI-attnWeights are other implementations of our estimated causal inference. The used model is Qwen2-VL-2B \cite{wang2024qwen2} on the COCO Caption dataset \cite{chen2015microsoft}.}
\label{fig_sup_vis_eci}
\end{figure*}

We have validated the effectiveness of the proposed Estimated Causal Inference (ECI) in Table \ref{tab_effec}. In this section, we present visualization results that illustrate how our ECI outperforms existing methods and alternative implementations, as shown in Fig. \ref{fig_sup_vis_eci}. The first baseline we consider is feature surgery \cite{li2025closer}, which is designed to mitigate redundant features along the class dimension. However, the challenge with multi-language models (MLLMs) lies in the correlated activations along the token prediction dimension, which is fundamentally different. As a result, feature surgery performs significantly worse than our ECI. Given the limited methods addressing correlated activations, we introduce additional baselines derived from other implementations of ECI: ECI-mean and ECI-attnWeights, as details in Supp. \ref{sec_supp_baselines}. Although these suboptimal implementations outperform feature surgery, they still yield inferior results compared to the final ECI. Notably, our ECI demonstrates superior performance in handling function words, producing significantly fewer redundant activations while achieving better recall of target objects. These results indicate that our ECI is well-designed and effective for mitigating correlated activations among the generated tokens of MLLMs.

\begin{table*}[h]
\centering
\footnotesize
\setlength\tabcolsep{7pt}
\begin{tabular}{ccccccccccc}
\hline 
\multirow{2}{*}{\textbf{Method}} & \multirow{2}{*}{\textbf{MLLM}} & \multicolumn{3}{|c|}{\textbf{COCO Caption}} &\multicolumn{3}{c|}{\textbf{GranDf}} & \multicolumn{3}{c}{\textbf{OpenPSG}} \\
 & & \multicolumn{1}{|c}{Obj-IoU} & Func-IoU & \textbf{F1-IoU} & \multicolumn{1}{|c}{Obj-IoU} & Func-IoU & \textbf{F1-IoU} & \multicolumn{1}{|c}{Obj-IoU} & Func-IoU & \textbf{F1-IoU} \\
\hline
CAM & \multirow{2}{*}{LLaVA1\_5-7B \cite{liu2024visual}} & 23.17 & 43.16 & 30.16 & 20.07 & 47.48 & 28.21 & 25.11 & 51.55 & 33.77 \\
TAM & & 27.65 & 61.43 & \textbf{38.13} & 20.71 & 59.15 & \textbf{30.68} & 28.57 & 61.06 & \textbf{38.93} \\
\cdashline{1-11}[1pt/5pt]
CAM & \multirow{2}{*}{LLaVA1\_5-13B \cite{liu2024visual}} & 24.82 & 51.18 & 33.43 & 21.34 & 43.99 & 28.74 & 26.65 & 48.45 & 34.39 \\
TAM & & 29.12 & 58.5 & \textbf{38.88} & 22.1 & 51.02 & \textbf{30.84} & 30.88 & 59.96 & \textbf{40.76} \\
\cdashline{1-11}[1pt/5pt]
CAM & \multirow{2}{*}{InternVL2\_5-2B \cite{chen2024internvl}} & 15.94 & 45.62 & 23.63 & 18.28 & 37.64 & 24.61 & 19.76 & 46.42 & 27.72 \\
TAM & & 21.38 & 65.1 & \textbf{32.19} & 20.48 & 85.93 & \textbf{33.08} & 23.0 & 86.86 & \textbf{36.36} \\
\cdashline{1-11}[1pt/5pt]
CAM & \multirow{2}{*}{InternVL2\_5-4B \cite{chen2024internvl}} & 18.23 & 40.95 & 25.23 & 20.91 & 44.52 & 28.46 & 21.28 & 34.7 & 26.38 \\
TAM & & 21.76 & 63.12 & \textbf{32.36} & 22.53 & 89.71 & \textbf{36.02} & 23.49 & 89.75 & \textbf{37.23} \\
\cdashline{1-11}[1pt/5pt]
CAM & \multirow{2}{*}{InternVL2\_5-8B \cite{chen2024internvl}} & 14.59 & 64.41 & 23.8 & 18.04 & 57.42 & 27.45 & 18.46 & 62.21 & 28.47 \\
TAM & & 19.98 & 66.53 & \textbf{30.73} & 21.56 & 85.95 & \textbf{34.47} & 21.73 & 88.74 & \textbf{34.91} \\
\cdashline{1-11}[1pt/5pt]
CAM & \multirow{2}{*}{Qwen2-VL-2B \cite{wang2024qwen2}} & 21.23 & 51.93 & 30.14 & 17.85 & 62.15 & 27.74 & 22.93 & 48.5 & 31.15 \\
TAM & & 27.37 & 68.44 & \textbf{39.1} & 18.65 & 88.97 & \textbf{30.83} & 26.26 & 92.99 & \textbf{40.95} \\
\cdashline{1-11}[1pt/5pt]
CAM & \multirow{2}{*}{Qwen2-VL-7B \cite{wang2024qwen2}} & 22.51 & 42.44 & 29.42 & 18.6 & 68.03 & 29.21 & 23.41 & 42.94 & 30.3 \\
TAM & & 28.13 & 71.85 & \textbf{40.43} & 19.88 & 90.57 & \textbf{32.61} & 26.94 & 89.88 & \textbf{41.45} \\
\hline 
\end{tabular}
\caption{\label{tab_sup_mllm} TAM shows wide applicability on diverse MLLMs and datasets beyond the CAM \cite{zhou2016learning} for all the experiments on the major F1-IoU (\%) metric at large margins. TAM can be used as a visual comparison approach, where Qwen2-VL models \cite{wang2024qwen2} show better visual explainability than LLaVA1\_6 \cite{liu2024visual} and InternVL2\_5 \cite{chen2024internvl} on the COCO Caption \cite{chen2015microsoft} and OpenPSG \cite{zhou2024openpsg} datasets.}
\end{table*}

\section{Visual Comparison Among Denoise Filters}
\label{sec:sup_vis_denoise}
Image denoising remains a traditional research topic,  but it is the first time to be introduced in the visual explanation field. The issue of noise has been addressed in Sec. \ref{sec:denoise}, where various methods aimed at noise reduction in transformers are discussed. However, residual noise persists even after these methods are applied. Consequently, it is essential to introduce denoising filters as a straightforward yet effective solution. Unlike conventional models that produce very small output sizes (e.g., $7 \times 7$), the output size of MLLMs is comparatively larger (e.g., $36 \times 36$). As a result, scatter-shaped noise is more likely to occur in MLLMs.

\begin{figure*}[h]
\centering
 \includegraphics[width=1\textwidth]{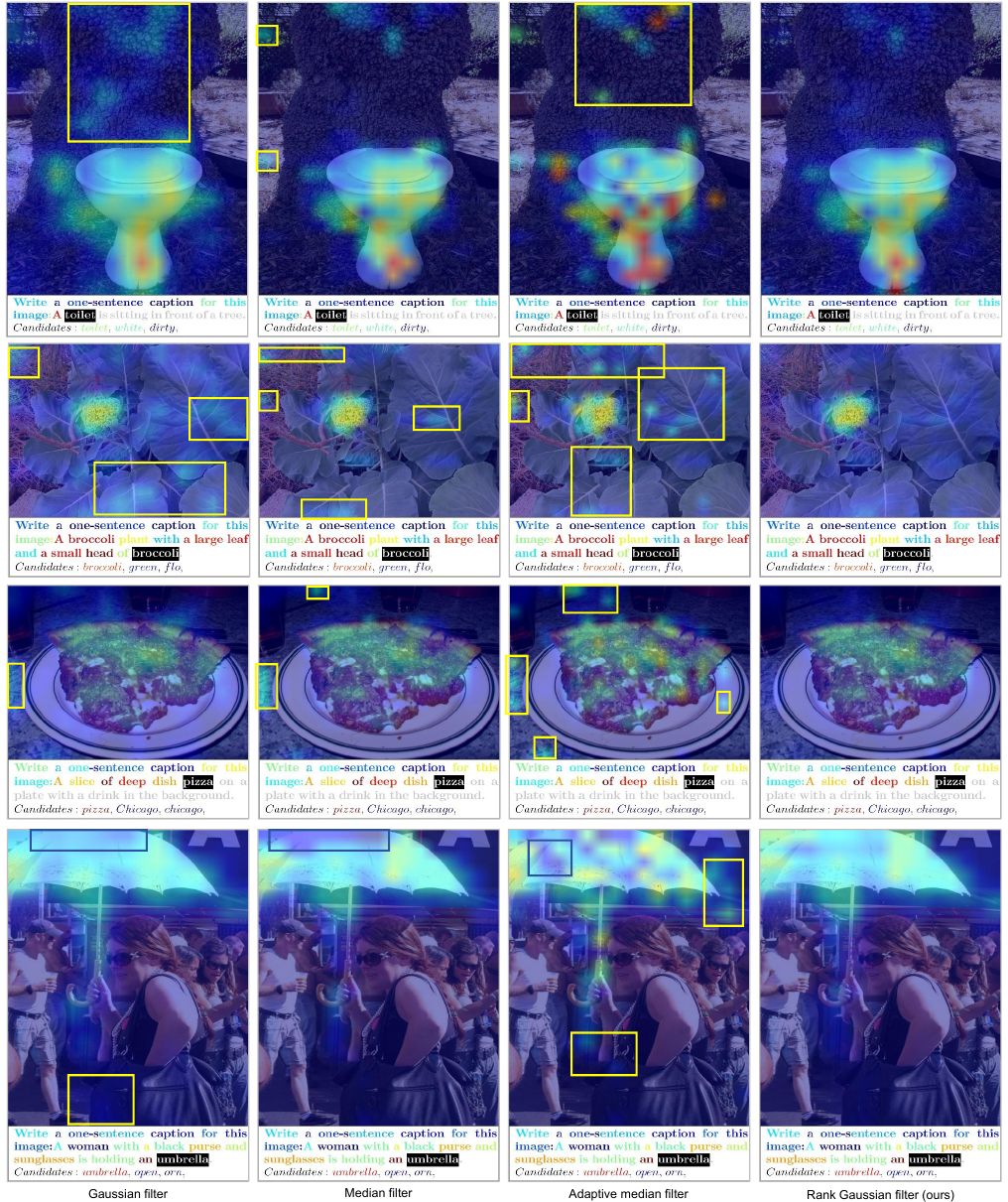}
\caption{\textbf{The proposed rank Gaussian filter is more effective than existing methods}. The yellow boxes indicate insufficient denoising, and the blue boxes mean over-denosing. The used model is Qwen2-VL-2B \cite{wang2024qwen2} on the COCO Caption dataset \cite{chen2015microsoft}.}
\label{fig_sup_vis_denoise}
\end{figure*}

These noises belong to the salt-and-pepper noises in general, which can be effectively addressed using median and Gaussian filters. While these methods do not represent the optimal solution, as illustrated in Fig. \ref{fig_sup_vis_denoise}. Specifically, the Gaussian filter proves inadequate in mitigating clustered noise, leaving many noises visible in the yellow boxes. The median filter reduces noise effectively, yet it still leaves behind unsolved scatter noise, with additional missing regions indicated by blue boxes in the final row. Similarly, the adaptive median filter \cite{chang2008adaptive} exhibits significant scatter noise, particularly near image edges where noise concentration is higher. In contrast, our proposed rank Gaussian filter demonstrates superior performance by amalgamating the strengths of both Gaussian and median filters, along with the novel technical enhancements discussed in Sec. \ref{sec:denoise}.

\section{Visualization of Ablation Study}
\label{sec:sup_vis_ablation}
\begin{figure*}[h]
\centering
 \includegraphics[width=0.97\textwidth]{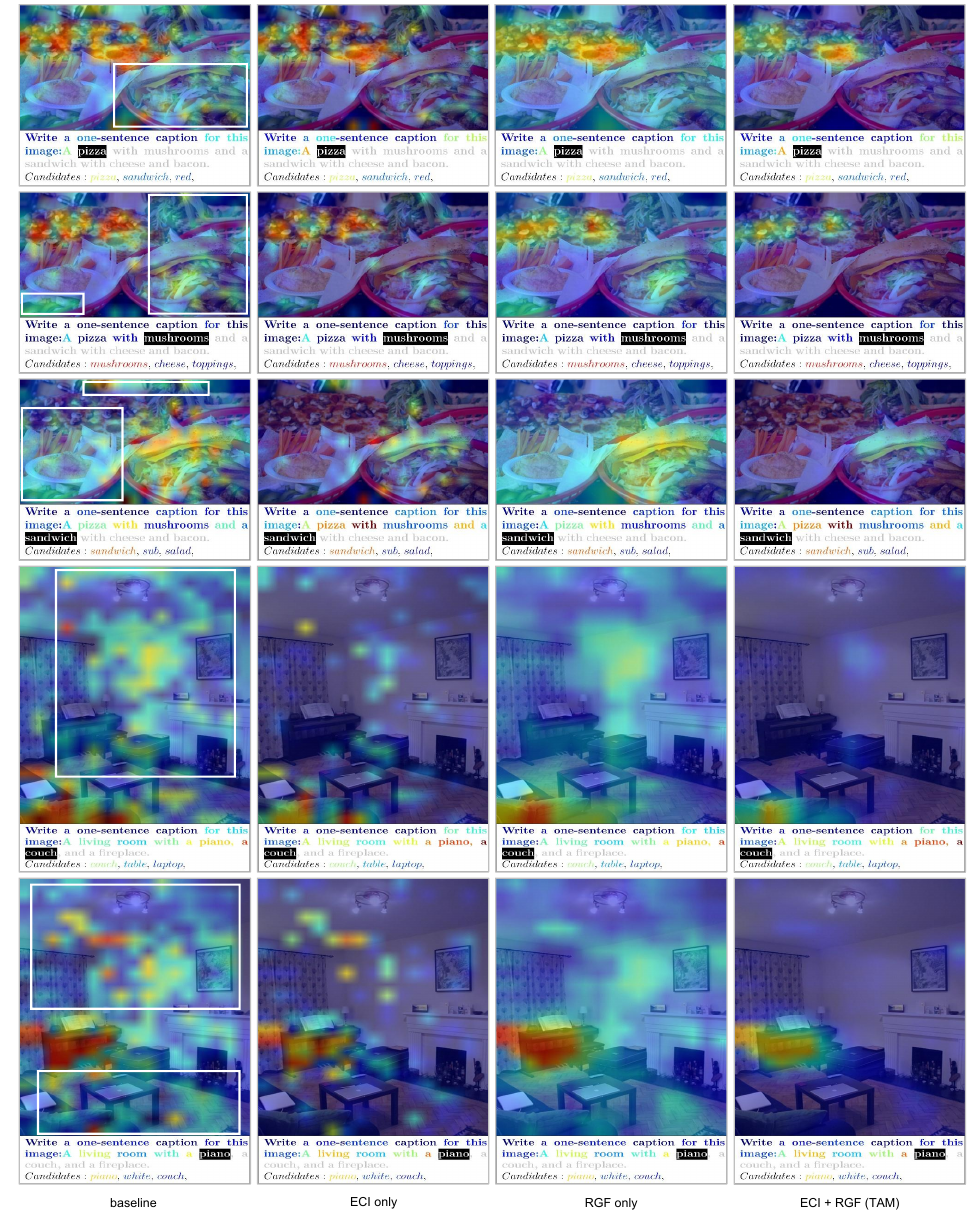}
\caption{\textbf{TAM involves two mutually beneficial modules.} (ECI only) The estimated causal inference can mitigate most of the interference activations, while some stubborn noise remains. (RGF only) The rank Gaussian filter can remove these noises, while redundant activations are obvious. (ECI + RGF) TAM integrates the advantages of both and achieves good explanation results far beyond the baseline \cite{zhou2016learning,selvaraju2017grad}. The white boxes indicate correlated activations solved by our method. The used model is Qwen2-VL-2B \cite{wang2024qwen2} on the COCO Caption dataset \cite{chen2015microsoft}.}
\label{fig_sup_vis_ablation}
\end{figure*}

We conducted ablation studies in Table \ref{tab_ablation}. In addition to the quantitative results, we present further visualizations in Fig. \ref{fig_sup_vis_ablation} to elucidate the effectiveness of these modules. The first column showcases the baseline method, CAM \cite{zhou2016learning} / Grad-CAM \cite{selvaraju2017grad}, which displays numerous redundant activations accompanied by noise, highlighted in white boxes. The proposed estimated causal inference (ECI) method in the second column effectively mitigates most correlated activations, although some persistent noise remains. The rank Gaussian filter in the third column successfully removes this noise, but redundant activations are still evident. By integrating these two innovative techniques into the proposed TAM, we achieve substantial explanatory results that leverage the strengths of both approaches. These examples illustrate the mutual benefits of the modules, leading to an overall improvement that exceeds the sum of their individual contributions, as shown in Table \ref{tab_ablation}.

\section{Explainability Results on Diverse MLLMs}
\label{sec_supp_mllms}

In addition to the explainability improvements highlighted in Table \ref{tab_mllm}, we present specific results in Table \ref{tab_sup_mllm}. This table reveals that the overall F1-IoU of the TAM ranges from 30.68\% to 41.45\% across three datasets and seven MLLMs. In contrast, the baseline CAM \cite{zhou2016learning}, which is considered a SoTA method in terms of performance and practicality (as shown in Table \ref{tab_sota}), achieves F1-IoU results ranging from 23.63\% to 34.39\%. These results clearly indicate that TAM demonstrates broader applicability and enhanced explainability across diverse MLLMs.

Moreover, TAM offers a unique perspective on evaluating MLLMs from an explainability standpoint, beyond existing metrics. For instance, LLaVA models \cite{liu2024visual} and Qwen2-VL models \cite{wang2024qwen2} exhibit higher F1-IoU scores than InternVL models \cite{chen2024internvl} on the COCO Caption \cite{chen2015microsoft} and OpenPSG \cite{zhou2024openpsg} datasets, while InternVL models excel on the GranDf dataset \cite{rasheed2024glamm}.

\begin{figure}[h]
\centering
 \includegraphics[width=8.2cm]{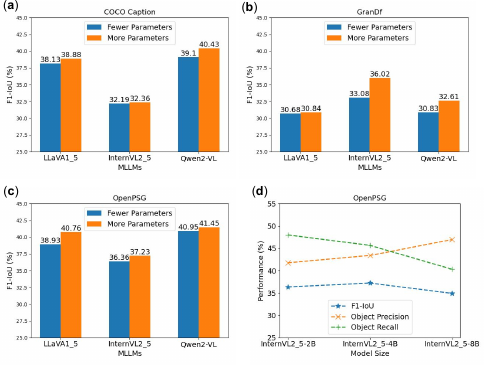}
\caption{Relation between model size and explainability. (\textbf{a}-\textbf{c}) F1-IoU (\%) results on LLaVA1\_5 \cite{liu2024visual} (7B, 13B), InternVL2\_5 \cite{chen2024internvl} (2B, 4B), and Qwen2-VL \cite{wang2024qwen2} (2B, 7B) across diverse datasets \cite{chen2015microsoft,rasheed2024glamm,zhou2024openpsg} indicates the explainability is increased with more model parameters within a certain range. (\textbf{d}) Increasing the parameters of InternVL2\_5 \cite{chen2024internvl} on the OpenPSG dataset \cite{zhou2024openpsg} improves object precision; however, this comes at the cost of decreased recall, which may negatively impact the F1-IoU score when the recall is too low.}
\label{fig_sup_size_effect}
\end{figure}

We also observe scalability in explainability across certain model sizes, as shown in Fig. \ref{fig_sup_size_effect}a-c. For example, LLaVA models show improvements from 7B to 13B, InternVL models from 2B to 4B, and Qwen2-VL models from 2B to 7B. This trend suggests a positive correlation between the scalability and explainability of MLLMs to a certain extent. When the model size are larger, the model tends to encode objects with fewer tokens, leading to a decrease in recall (see Fig. \ref{fig_sup_size_effect}d) and an increase in precision. Subsequently, the Obj-IoU decreases due to a more significant decrease in recall.

\section{TAM for MLLM Visual Comparison}
\label{sec:sup_vis_mllms}

Generally, researchers MLLMs using quantitative metrics or textual outputs, while visual comparisons remain underexplored. In contrast, visual evaluations are prevalent in conventional models such as Convolutional Neural Networks (CNNs) and Vision Transformers (ViTs), which effectively illustrate the enhanced representational capabilities of new models. The absence of a dedicated explanation tool for MLLMs may contribute to this gap. The proposed TAM addresses this limitation, enabling researchers to conduct visual comparisons of their MLLMs against existing models, beyond the conventional textual comparison.

\begin{figure*}[h]
\centering
 \includegraphics[width=1\textwidth]{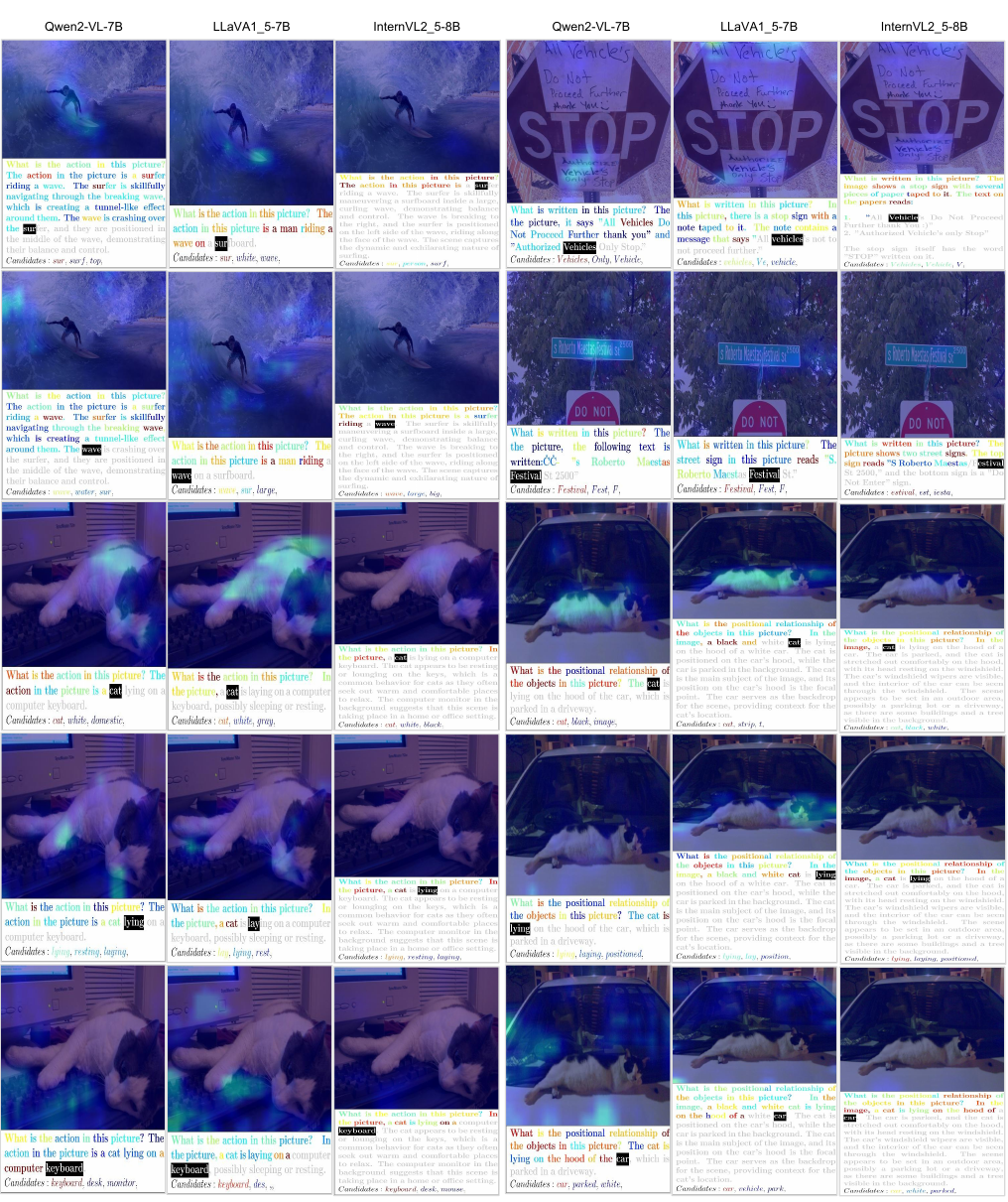}
\caption{\textbf{TAM supports visual comparison among MLLMs about attributes}. Qwen2-VL-7B \cite{wang2024qwen2} presents good visual explainability beyond LLaVA1\_5-7B \cite{liu2024visual} on texts. InternVL2\_5-8B \cite{chen2024internvl} focuses on textual content with more red texts and weaker visual activations.}
\label{fig_sup_vis_mllms_attri}
\end{figure*}

\begin{figure*}[h]
\centering
 \includegraphics[width=1\textwidth]{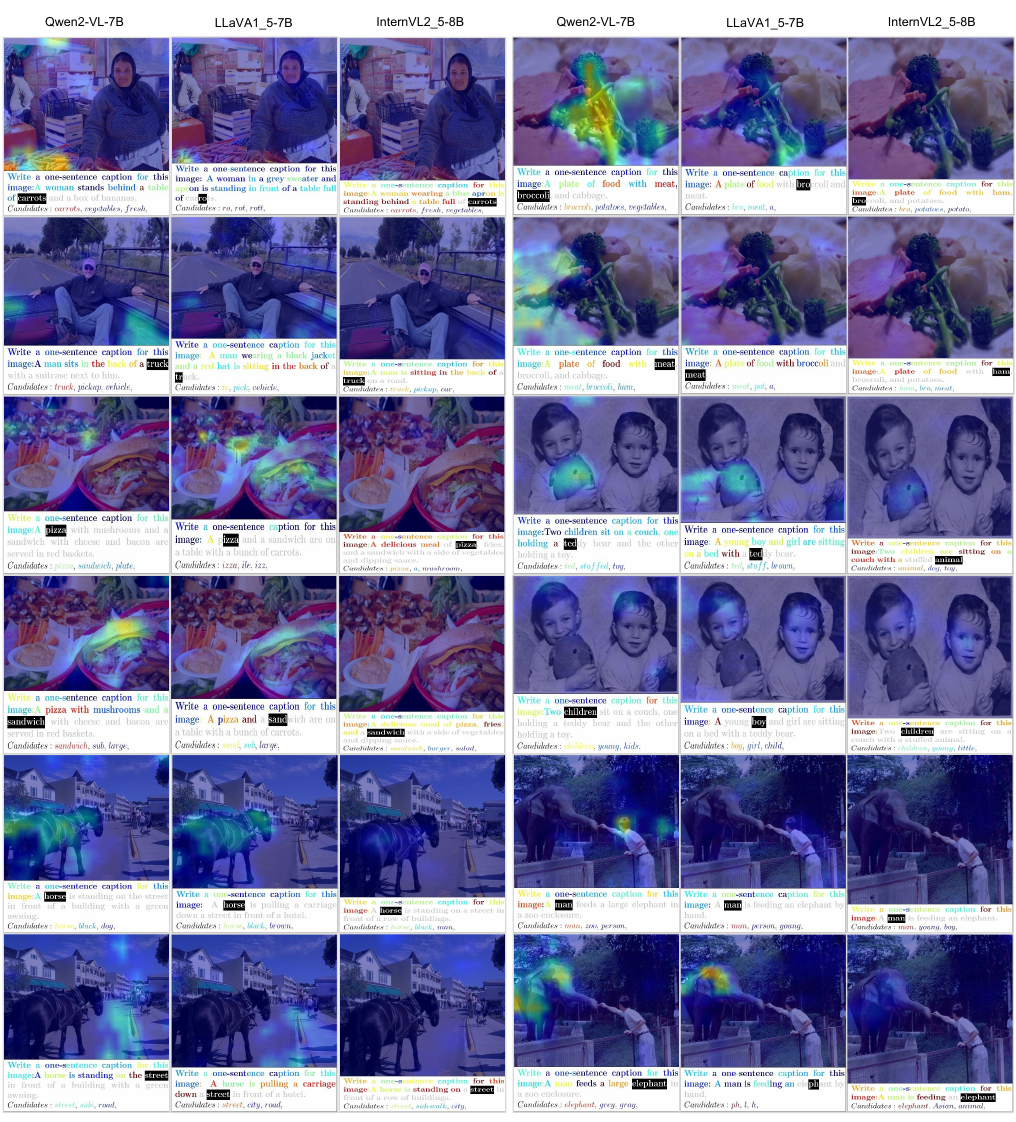}
\caption{\textbf{TAM supports visual comparison among MLLMs about objects} on the COCO Caption \cite{chen2015microsoft} dataset. Qwen2-VL-7B \cite{wang2024qwen2} presents the best visual results with less correlation (e.g., pizza vs. sandwich in the third row) and shows a stronger activation degree.}
\label{fig_sup_vis_mllms_coco}
\end{figure*}

We present visual comparisons among MLLMs in Fig. \ref{fig_sup_vis_mllms_attri}, focusing on attributes, and Fig. \ref{fig_sup_vis_mllms_coco}, which highlights object recognition. One key finding is that Qwen2-VL-7B \cite{wang2024qwen2} surpasses LLaVA1\_5-7B \cite{liu2024visual} in text attributes, as illustrated in Fig. \ref{fig_sup_vis_mllms_attri}. Furthermore, it demonstrates superior performance in the object cases compared with LLaVA. For instance, Qwen2-VL-7B exhibits stronger activations in the top two rows of Fig. \ref{fig_sup_vis_mllms_coco} and lacks correlation with the sandwich when interpreting the pizza in the third row.

In contrast, InternVL2\_5-8B \cite{chen2024internvl} tends to overemphasize textual information, resulting in weaker visual responses. Although it generates highly detailed textual content that exceeds the length of LLaVA's outputs, its activation map displays comparatively lower intensity. This is evidenced by the pronounced responses in textual areas (illustrated by the increased red regions). Despite reasonable activations in several successful cases (indicated by the light blue regions), InternVL2\_5-8B exhibits lower quantitative performance, as shown in Table \ref{tab_sup_mllm}.

These visual findings underscore that TAM offers a novel perspective for comparing MLLMs, providing deeper insights, particularly when quantitative results are closely matched across certain datasets. Researchers can leverage visual comparisons to highlight the advantages of their models.

\section{Extensive Cases about Attributes Analysis}
\label{sec:sup_attributes}

\begin{figure*}[h]
\centering
 \includegraphics[width=1\textwidth]{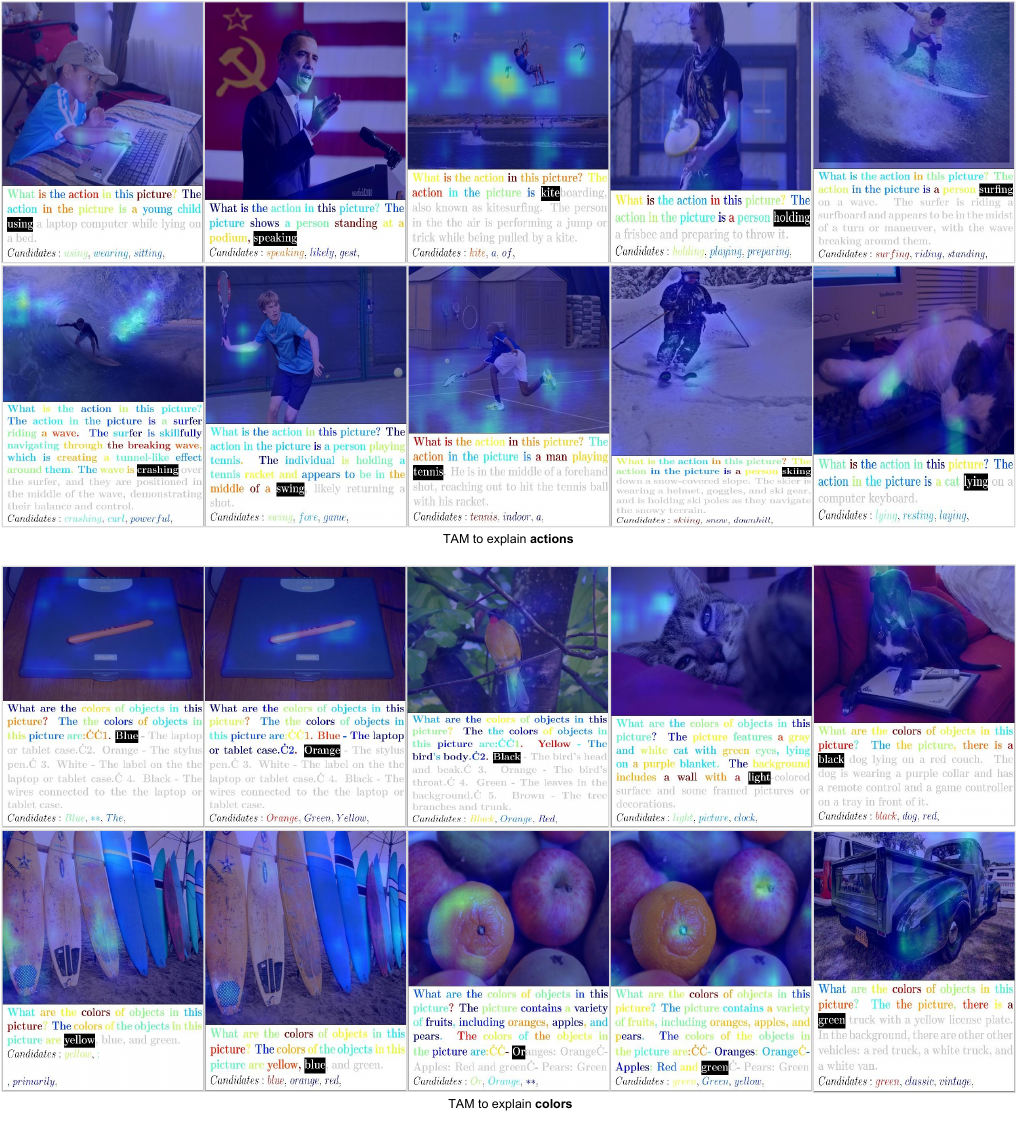}
\caption{\textbf{TAM supports explaining attributes of MLLMs at high-quality} for the Qwen2-VL-7B \cite{wang2024qwen2} about action and colors.}
\label{fig_sup_vis_attri1}
\end{figure*}

\begin{figure*}[h]
\centering
 \includegraphics[width=1\textwidth]{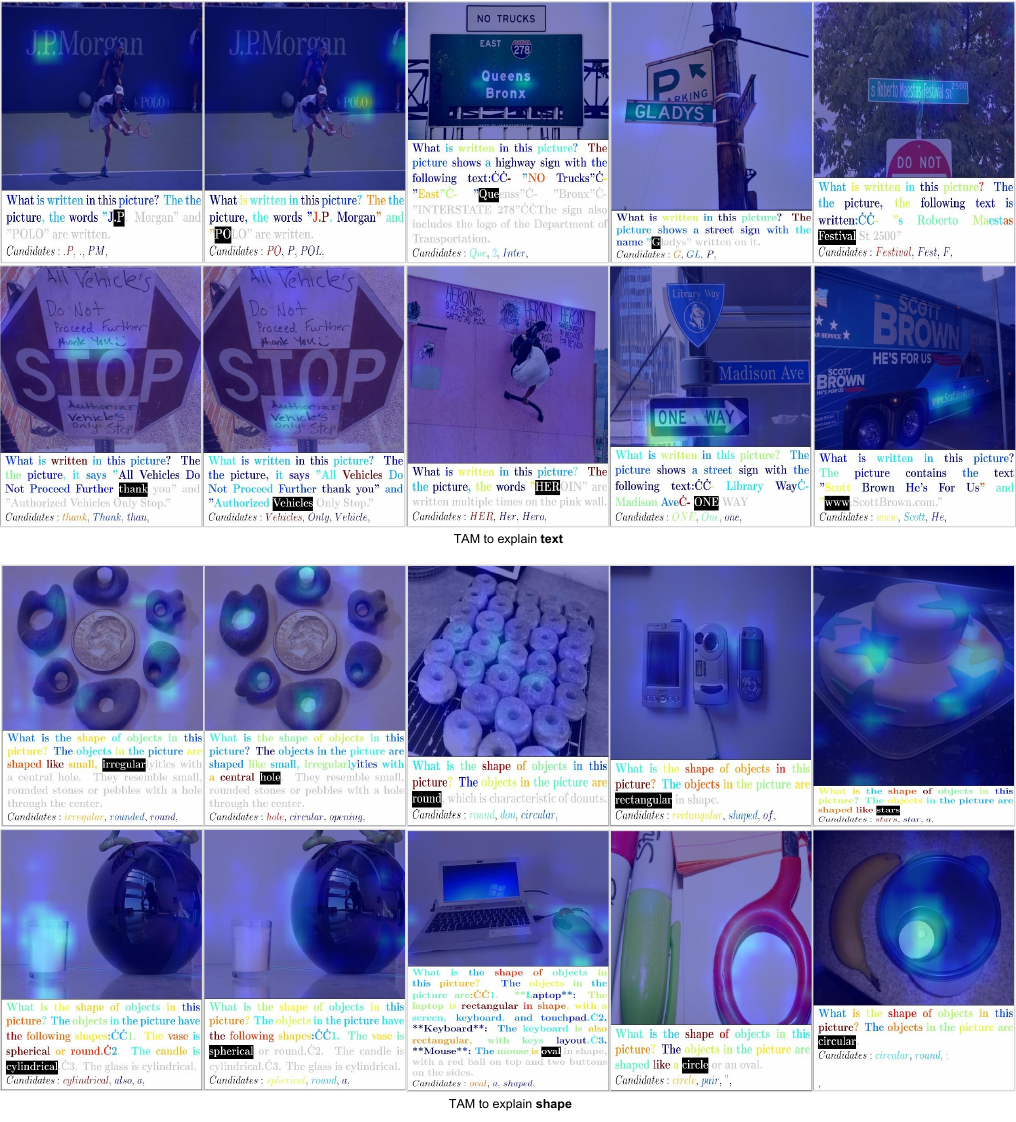}
\caption{\textbf{TAM supports explaining attributes of MLLMs at high-quality} for the Qwen2-VL-7B \cite{wang2024qwen2} about text and shape.}
\label{fig_sup_vis_attri2}

\end{figure*}

\begin{figure*}[h]
\centering
 \includegraphics[width=1\textwidth]{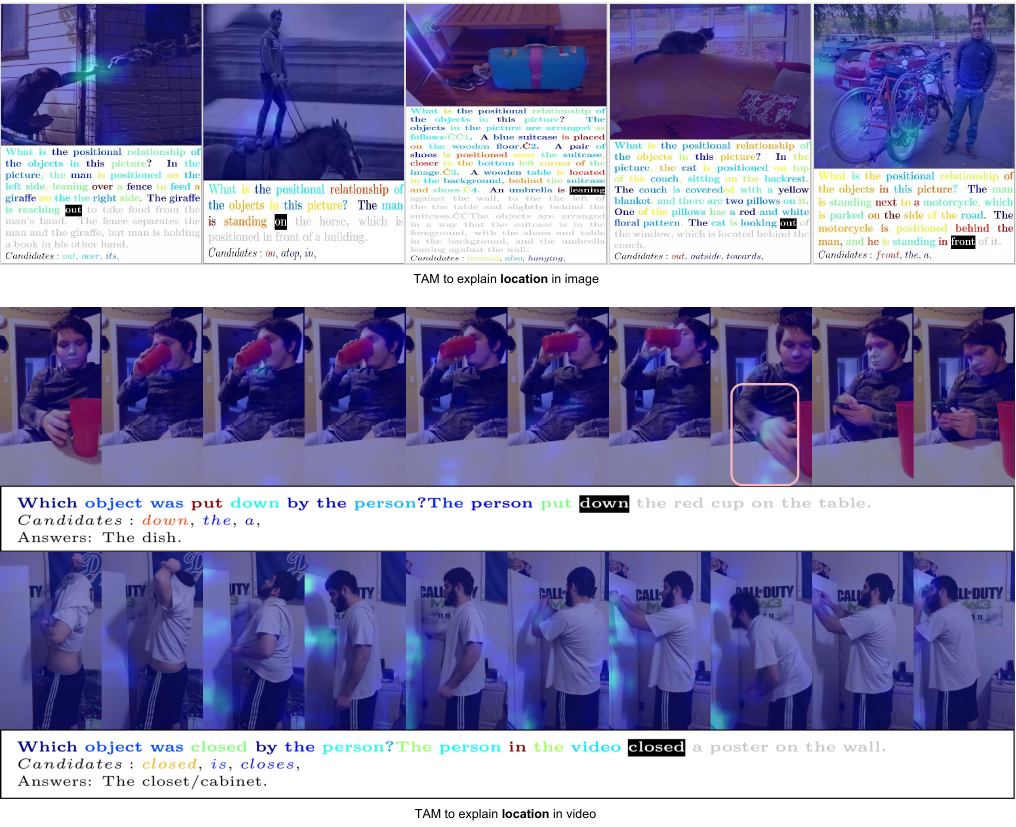}
\caption{\textbf{TAM supports explaining attributes of MLLMs at high-quality for both images and videos.} The images are processed by the Qwen2-VL-7B \cite{wang2024qwen2} from the COCO Caption dataset \cite{chen2015microsoft} and we use the Qwen2-VL-2B for videos from the STAR dataset \cite{wu2star}.}
\label{fig_sup_vis_attri3}
\end{figure*}

The proposed TAM enables users to analyze the fine-grained attributes of MLLMs. These attributes contribute to a deeper understanding of how the model works. We present various visualizations of attributes, including actions and colors in Fig. \ref{fig_sup_vis_attri1}, text and shapes in Fig. \ref{fig_sup_vis_attri2}, and locations for both images and videos in Fig. \ref{fig_sup_vis_attri3}.

\begin{figure*}[h]
\centering
 \includegraphics[width=1\textwidth]{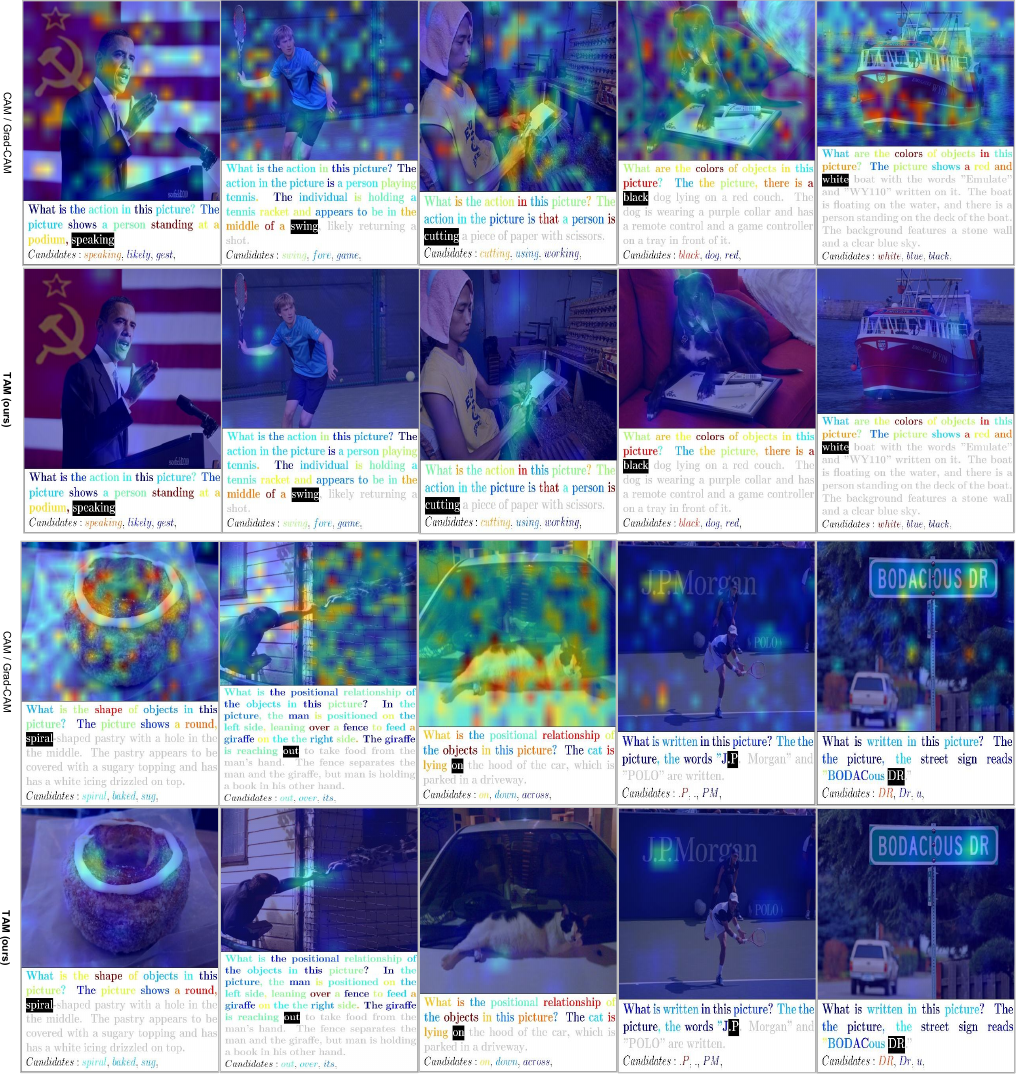}
\caption{\textbf{TAM exceeds existing methods on attribute explanation.} The images are processed by the Qwen2-VL-7B \cite{wang2024qwen2} from the COCO Caption dataset \cite{chen2015microsoft} compared with the baseline.  "CAM / Grad-CAM" indicates CAM \cite{zhou2016learning} and Grad-CAM \cite{selvaraju2017grad} are equivalent for MLLM, as discussed in Supp. \ref{sec_supp_baselines}.}
\label{fig_sup_vis_attri_comp}
\end{figure*}

The results indicate that the tested model, Qwen2-VL-7B \cite{wang2024qwen2}, possesses the capability to comprehend diverse attributes with a high degree of explainability. Furthermore, we compare existing methods \cite{zhou2016learning, selvaraju2017grad} with our proposed TAM in Fig. \ref{fig_sup_vis_attri_comp}, where our method demonstrates significantly superior explanation quality. These activation maps provide visual evidence for the generated content, thereby enhancing the model's credibility.

\section{TAM for Biased Scenario}
\label{sec:sup_bias}

\begin{figure}[H]
\centering
 \includegraphics{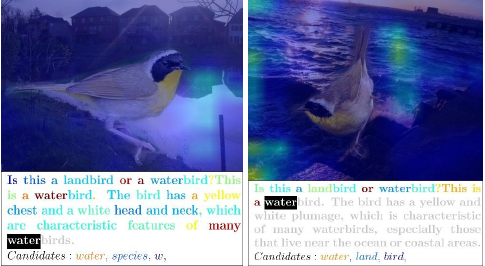}
\caption{TAM supports analyzing biased scenarios. The landbirds in these two images were mistakenly classified as waterbirds due to the synthesized water backgrounds. The TAM identified that this biased recognition arises from the influence of the background.}
\vspace{-0.4cm}
\label{fig_sup_vis_biased}
\end{figure}

The Task-Aware Mask (TAM) framework is capable of supporting the analysis of biased scenarios. In Fig. \ref{fig_sup_vis_biased}, we investigate whether the background environment unexpectedly influences the classification of target categories. The output text indicates that images of terrestrial birds with synthetic aquatic backgrounds were misclassified as waterbirds, suggesting a significant bias introduced by background features in the model's predictions. We conducted an in-depth analysis of this phenomenon using the TAM.

TAM effectively separates the contributions of different regions within the image to the classification decision, allowing for precise localization of the source of bias. Our research reveals that the synthetic aquatic background exerts a substantial influence on the model's internal representations, leading it to favor categorizing images as waterbirds. This finding underscores the importance of considering background information during the model training and evaluation processes. Over-reliance on background features rather than the characteristics of the target itself may result in systematic misjudgments in scenarios that include synthetic or artificially manipulated backgrounds. The TAM-based analysis provides an effective diagnostic tool for identifying issues like background bias.


\section{Extensive Failure Cases Study}
\label{sec:sup_failure}

\begin{figure*}[h]
\centering
 \includegraphics[width=1\textwidth]{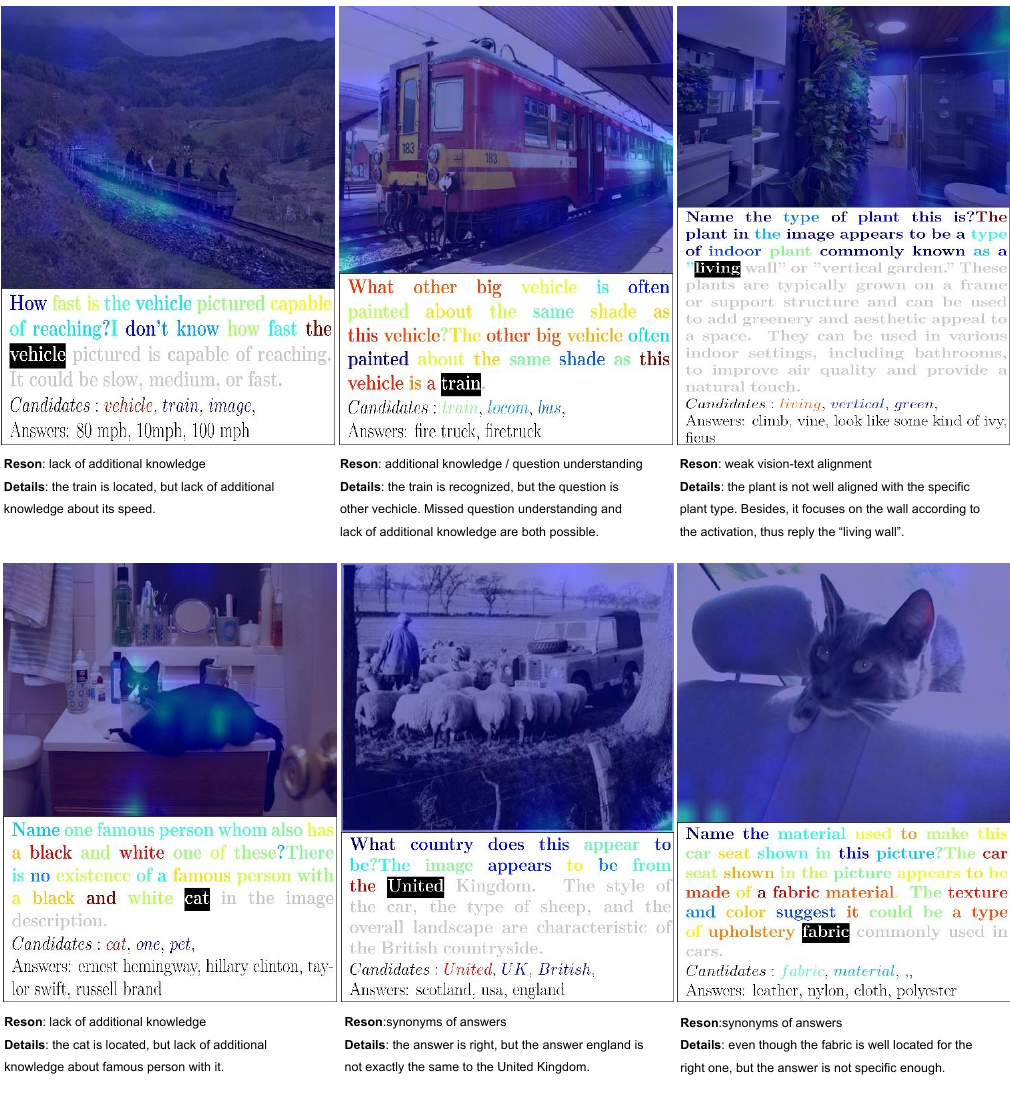}
\caption{\textbf{TAM supports failure case analysis for deeper understanding} with error reason and analysis details using Qwen2-VL-2B \cite{wang2024qwen2} on the QK-VQA dataset \cite{okvqa}.}
\label{fig_sup_vis_failure_case}
\end{figure*}

An important function of TAM is to support developers in analyzing failure cases, thereby deepening their understanding of the model's shortcomings and enabling the development of better MLLMs. Generally, developers analyze errors by comparing the reply and answer, while TAM provides a clear visual view to understand them with more insights. As shown in Fig. \ref{fig_sup_vis_failure_case}, we list several failure cases with the error reason and corresponding analysis. We find that sometimes the model can successfully locate the target object, but lacks additional knowledge related to it thus replying falsely or refusing to answer (e.g., the train and cat in the left of Fig. \ref{fig_sup_vis_failure_case}). If the model focus on other regions out of the target, the answer is possibly to be wrong. For example, we the model looks at the wall, it replies “living wall”, instead of the specific plant type the user asked for. Another error type is tolerable, that is synonyms, hypernyms, or hyponyms of answers (e.g., UK vs. England, fabric vs. nylon). 

\begin{figure*}[h]
\centering
 \includegraphics[width=1\textwidth]{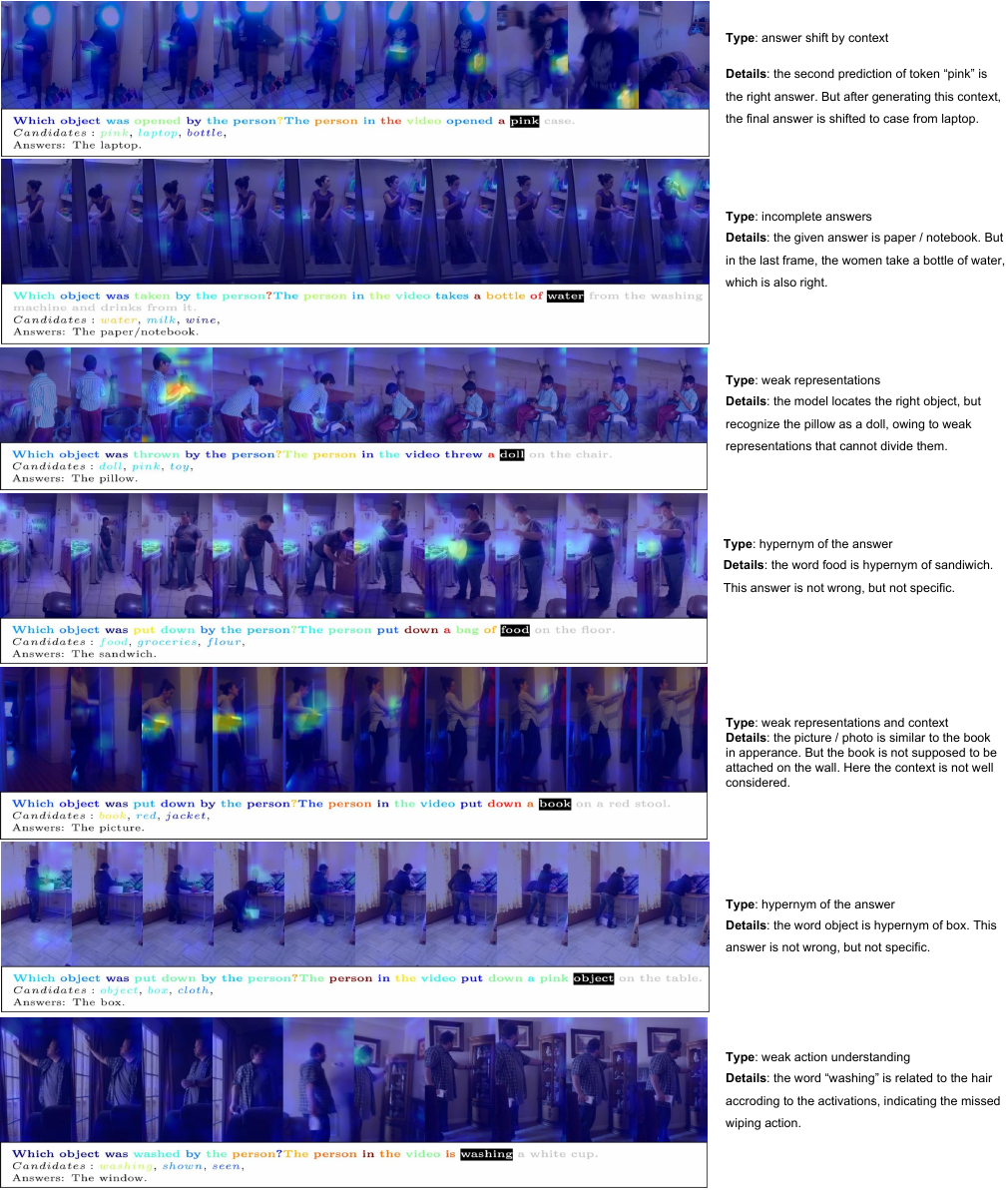}
\caption{\textbf{TAM supports failure case analysis for videos} with error reason and analysis details using Qwen2-VL-2B \cite{wang2024qwen2} on the STAR dataset \cite{wu2star}.}
\label{fig_sup_vis_video_failure_case}
\end{figure*}

We further conduct case analysis on videos using Qwen2-VL-2B \cite{wang2024qwen2} in Fig. \ref{fig_sup_vis_video_failure_case}. Some error types are interesting. In the first row, we find the model already knows the object is a laptop when generating the token “pink”. But it turns to the case sequentially. It indicates the answer is shifted by context (maybe trained with some corpus including “pink case”). Besides, the representation is not strong enough, and the model cannot divide the pattern of the pillow and doll in the third row. In the fifth row, the picture with a border is similar to a book, while it is attached to the wall. From this context, we can know this is a picture instead of a book, indicating the weak capacity to integrate context. For the last row, the attention is located on the hair, suggesting the model predicts the “washing” according to the hair, instead of the window. All these examples prove that TAM can provide more cues and insights to analyze failure cases.

\section{Extensive Success VQA Examples}
\label{suc:sup_vis_vqa}
\begin{figure*}[h]
\centering
 \includegraphics[width=1\textwidth]{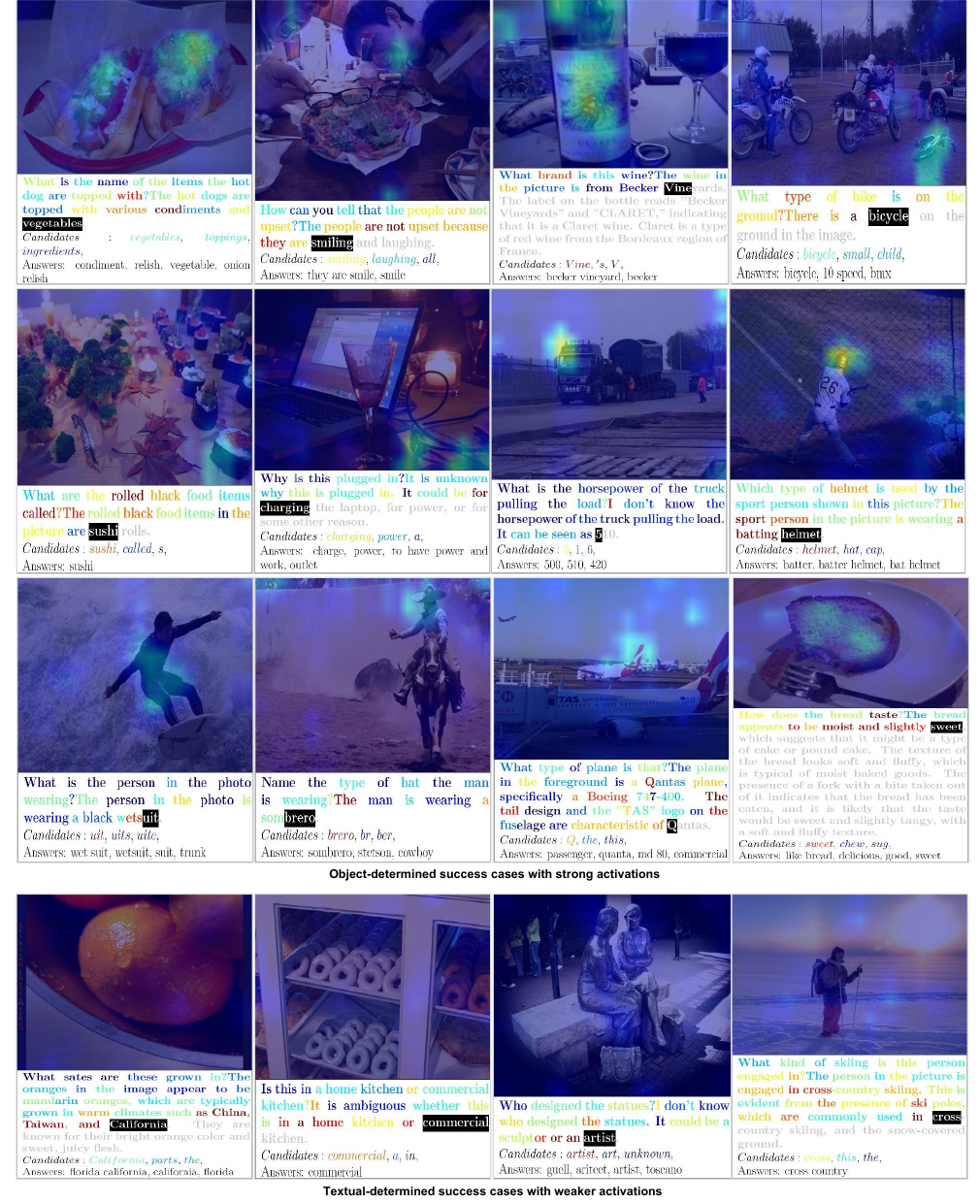}
\caption{\textbf{TAM presents good visual explanation result for the VQA dataset} with extensive successful examples on the QK-VQA dataset \cite{okvqa} using Qwen2-VL-2B \cite{wang2024qwen2}. These cases are dependent on different information, divided into “Object-determined” type and “Textual-determined” type, with higher and lower activation degrees, respectively.}
\label{fig_sup_vis_vqa}
\end{figure*}

In addition to the failure cases illustrated in Fig. \ref{fig_sup_vis_failure_case}, we present extensive success Visual Question Answering (VQA) examples in Fig. \ref{fig_sup_vis_vqa}. These visualization results indicate that the Token Activation Model (TAM) is applicable not only to caption-based datasets but also to VQA datasets, such as QK-VQA \cite{okvqa}. From the figure, we observe that certain images are well-aligned with the generated tokens, which include objects, actions, texts, and patterns (e.g., the Qantas logo), thereby facilitating accurate predictions. However, some cases are not primarily object-determined; they rely heavily on textual cues, as seen with terms like "commercial" and "cross" in the last row. This analysis allows us to discern the sources of predictions based on activation levels: higher responses indicate strong visual relevance, while lower responses suggest a greater reliance on textual information.

\section{Examples about Video Visualization}
\label{sec:sup_vis_video}
\begin{figure*}[h]
\centering
 \includegraphics[width=1\textwidth]{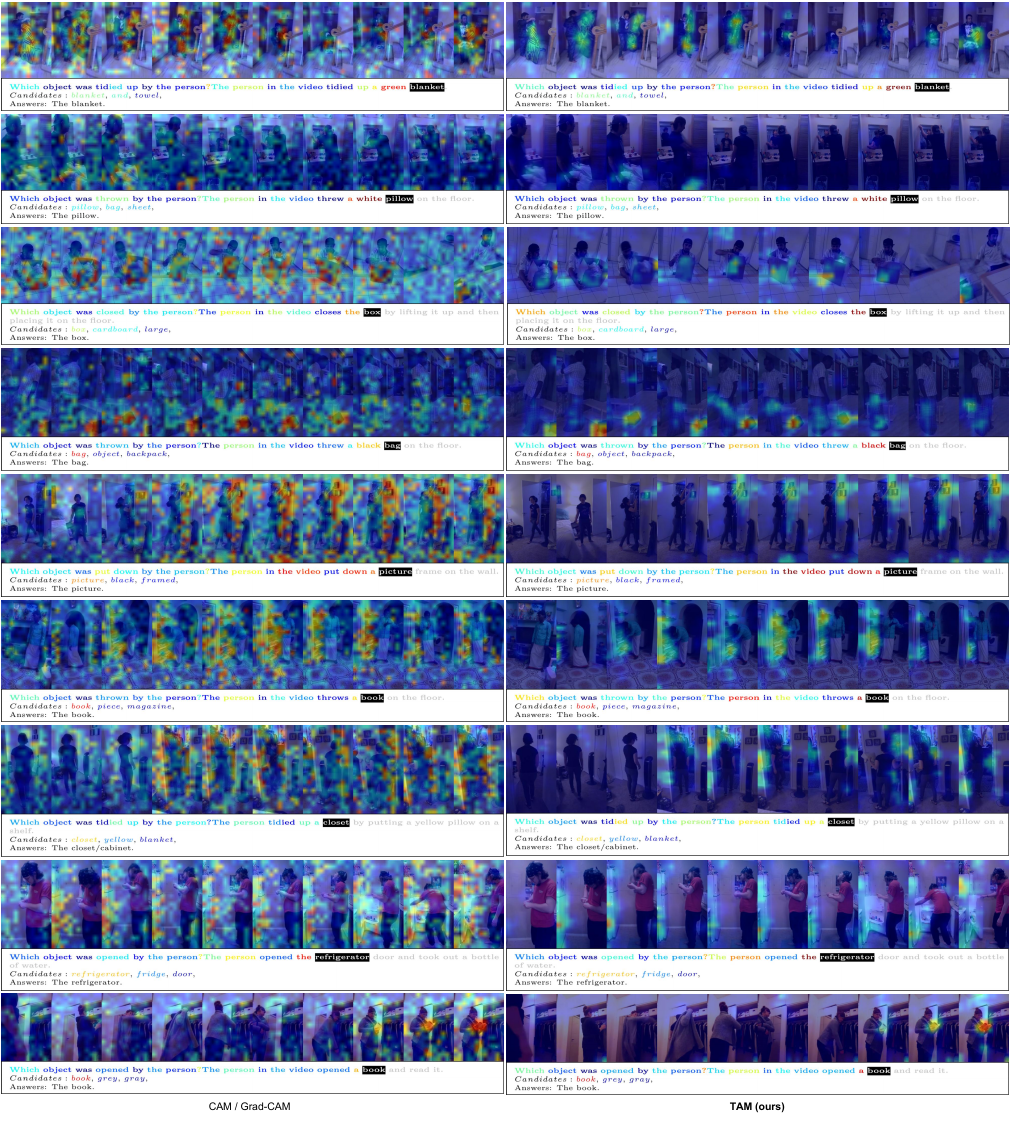}
\caption{Comparison about video visualization between our TAM and CAM \cite{zhou2016learning} / Grad-CAM \cite{selvaraju2017grad} (equivalent to MLLM) on the STAR dataset \cite{wu2star} using Qwen2-VL-2B \cite{wang2024qwen2}. \textbf{TAM presents much clearer visualization results with fewer redundant activations and noises}.}
\label{fig_sup_vis_videos}
\end{figure*}

Video modality is a crucial input type for MLLMs; however, it has seldom been studied in the explainability aspect. We compare our TAM with conventional methods \cite{zhou2016learning, selvaraju2017grad}, as illustrated in Fig. \ref{fig_sup_vis_videos}, using Qwen2-VL-2B \cite{wang2024qwen2} on the STAR dataset \cite{wu2star} for video understanding. It is evident that TAM produces significantly clearer video visualization results compared to CAM \cite{zhou2016learning} and Grad-CAM \cite{selvaraju2017grad}, both of which are well-established methods, as shown in Table \ref{tab_sota}. Specifically, TAM effectively reduces redundant activations and minimizes noise, allowing users to concentrate on target objects and observe the raw video more clearly. Additionally, we provide case studies in Fig. \ref{fig_sup_vis_video_failure_case} for video error analysis.

\section{Corner Case About Reasoning}
\label{sup_sec_reasoning}
\begin{figure*}[h]
\centering
 \includegraphics[width=1\textwidth]{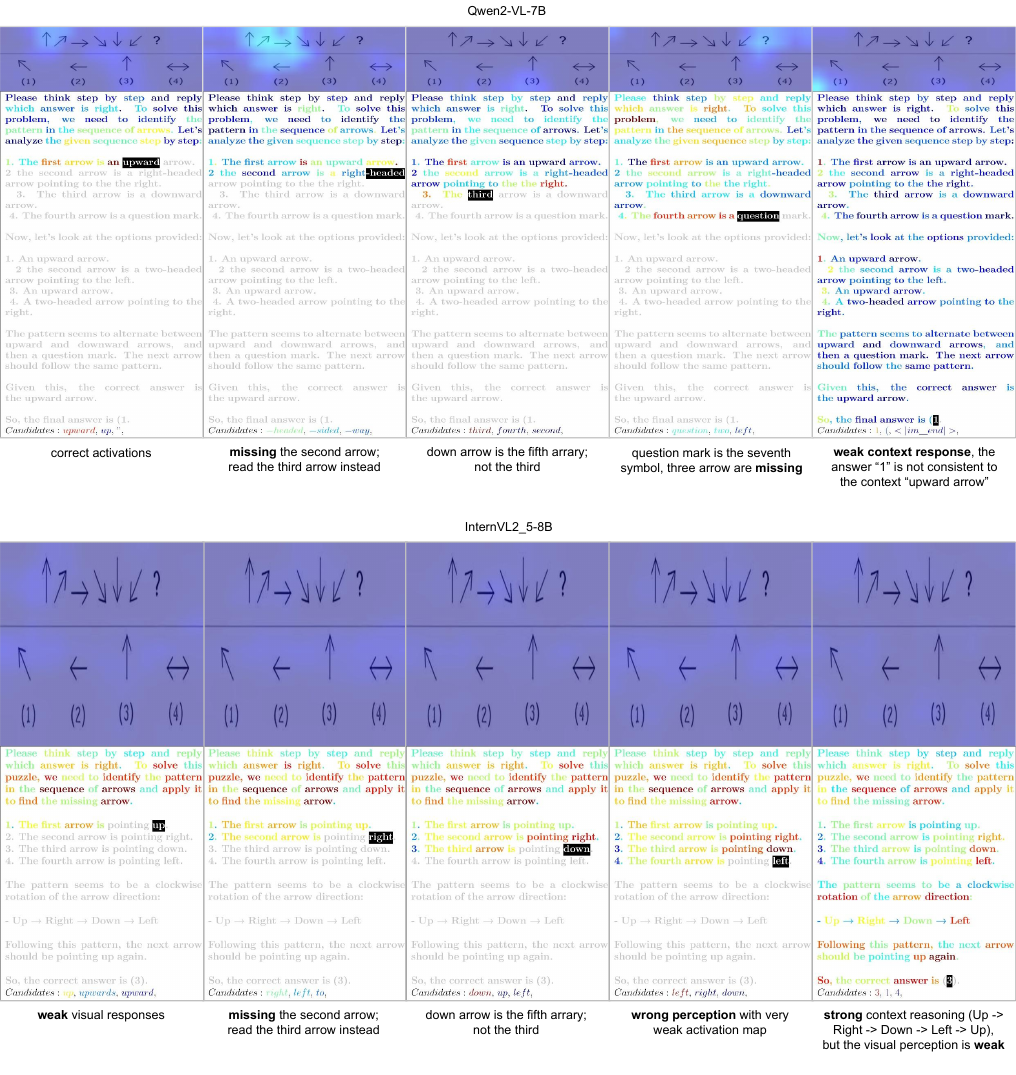}
\caption{\textbf{Visual reasoning corner case analyzed using TAM}. The analysis reveals that both Qwen2-VL-7B \cite{wang2024qwen2} and InternVL2\_5-8B \cite{chen2024internvl} incorrectly answered the question due to incomplete recognition of all the arrows. Missing arrows led to erroneous reasoning. Qwen2-VL-7B demonstrated weaker reasoning capabilities compared to InternVL2\_5-8B, exhibiting inconsistent context understanding and low text activations. In contrast, while InternVL2\_5-8B provided a stronger text response with correct reasoning logic, its visual perception and activation degree are weaker.}
\label{fig_sup_vis_reasoning}
\end{figure*}

TAM serves as a valuable tool for analyzing the visual reasoning processes of MLLM. In Fig. \ref{fig_sup_vis_reasoning}, we present a corner case of visual reasoning and analyze it using TAM. We find that both Qwen2-VL-7B \cite{wang2024qwen2} and InternVL2\_5-8B \cite{chen2024internvl} provided incorrect answers in this case. TAM reveals that the primary issue arises from incomplete recognition of all arrows in the problem, with missing arrows leading to subsequent reasoning errors. Furthermore, the reasoning capability of Qwen2-VL-7B is weaker than that of InternVL2\_5-8B in this instance, as evidenced by inconsistent context understanding and very low text activations. In contrast, while InternVL2\_5-8B exhibited stronger textual responses and correct logical reasoning, it compromised visual perception and interpretability, ultimately resulting in an incorrect answer as well. These analyses offer valuable insights for future model optimization and highlight the broad applicability of TAM.

\section{TAM for Multi-image Conversation}
\label{sec:sup_multi_img}

\begin{figure*}[h]
\centering
 \includegraphics[width=1\textwidth]{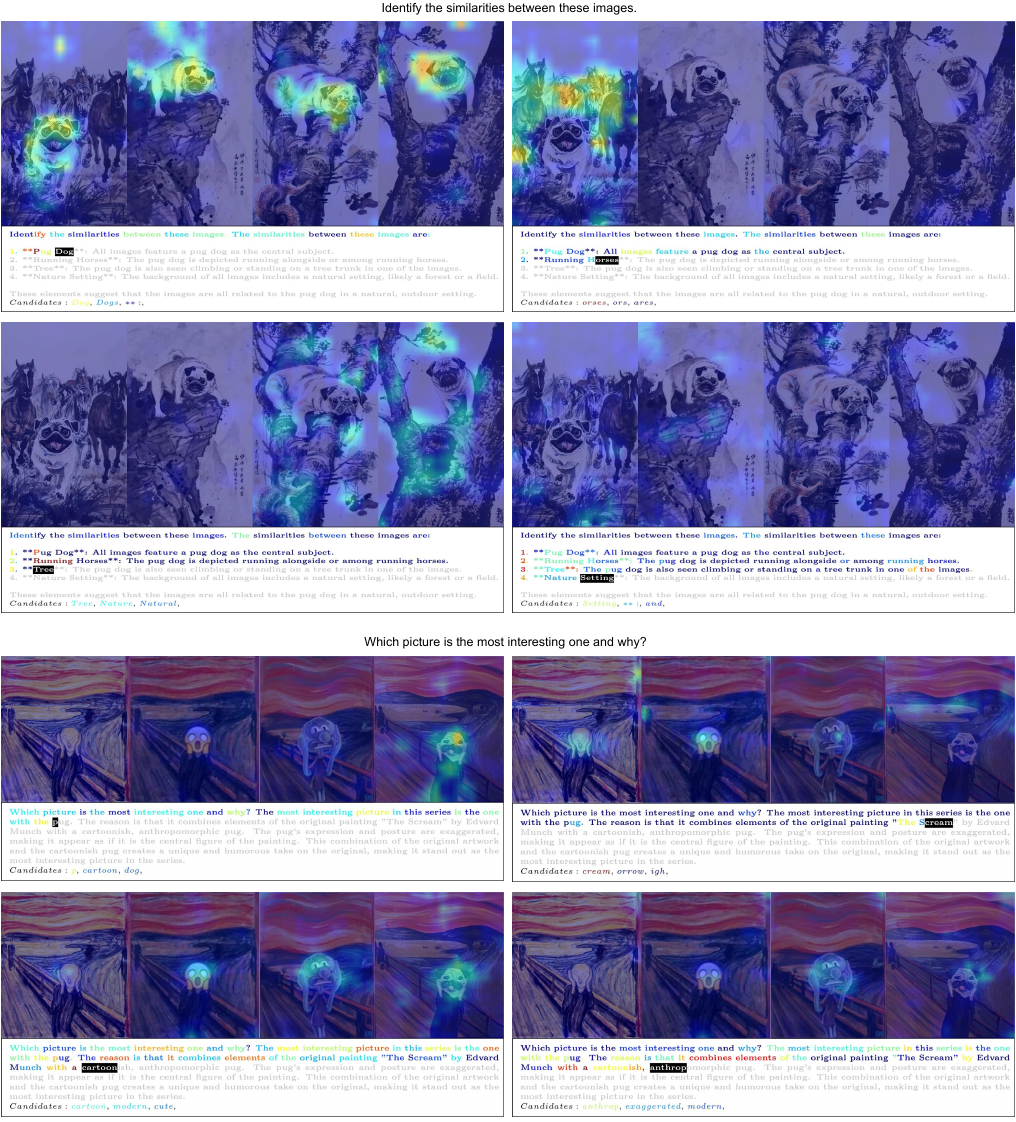}
\caption{\textbf{TAM supports multi-image conversation of MLLM showing wide applicability.} It generates high-quality activation maps for four input images in the first case using the Qwen2-VL-2B \cite{wang2024qwen2}. In the second case, TAM explains why the model regards the last image as the most interesting one, providing visual cues for key tokens. These results showcase the model's effectiveness in multi-image scenarios, highlighting its versatility in handling complex visual data.}
\label{fig_sup_vis_multi_img}
\end{figure*}

Conventional models generally have a single input and a single output, whereas the characteristic of MLLM is that it supports multiple inputs and multiple token outputs. We provide examples of multi-image conversation in this chapter, as shown in Fig. \ref{fig_sup_vis_multi_img}. In the first example, Qwen2-VL-2B \cite{wang2024qwen2} can identify the main elements from four different images, accompanied by high-quality explanation results provided by TAM. In the second example, TAM explains from a visual perspective why Qwen2-VL-2B considers the fourth image the most interesting. TAM presents detailed activation maps, explaining specific focus points on attributes such as object (pug), artwork (The Scream), image style (cartoon), and painting style (anthropomorphic). These results demonstrate the broad applicability of TAM, which can support various new capabilities of MLLMs.

\section{TAM for Multi-turn Conversation}
\label{sec:sup_multi_turn}

\begin{figure*}[h]
\centering
 \includegraphics[width=1\textwidth]{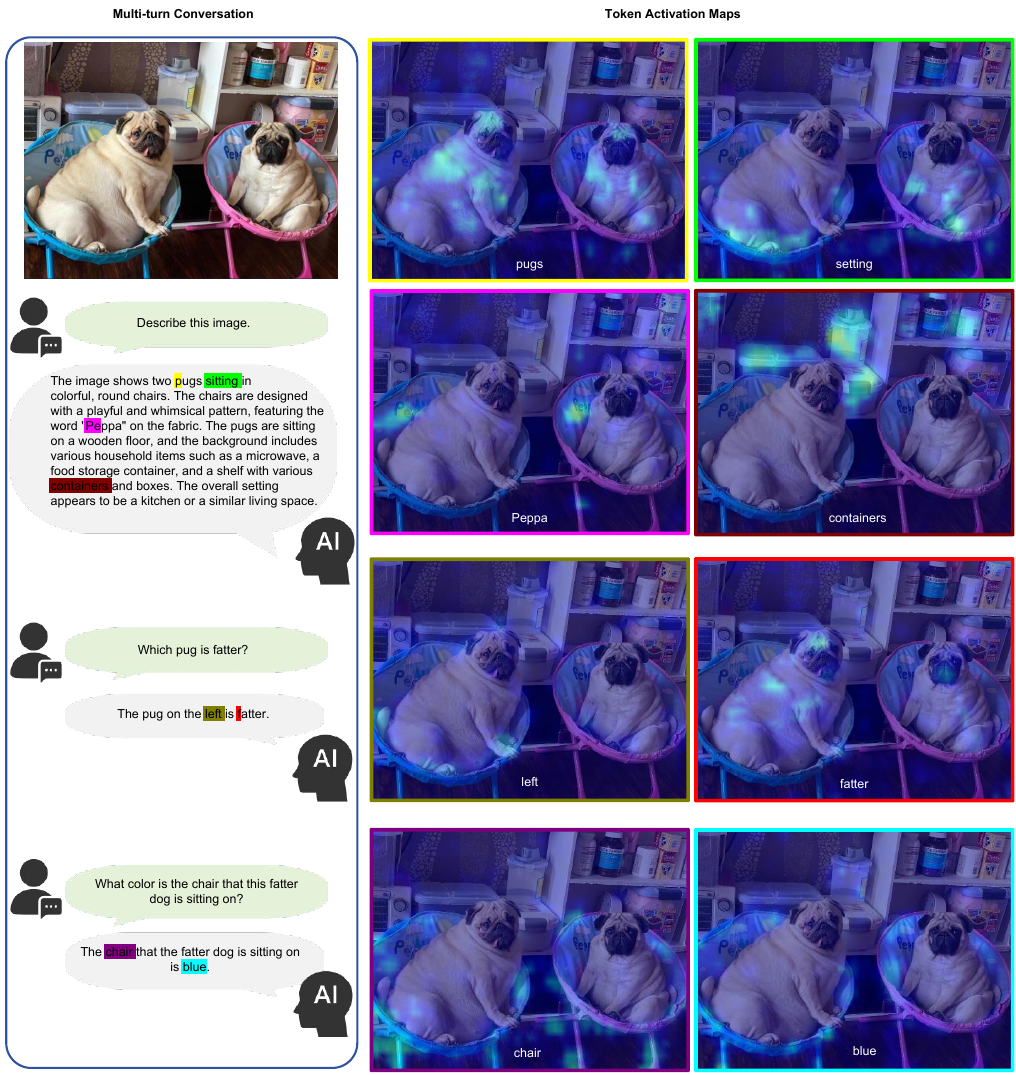}
\caption{\textbf{TAM supports multi-turn conversation of MLLM for diverse attributes}. TAM presents high-quality visual explanation results on Qwen2-VL-2B \cite{wang2024qwen2} regarding attributes such as objects, actions, and text in the first round. Then, the user inquires about the fatter dog and the color of a chair in the second and third rounds, respectively. Activation maps suggest TAM is capable of explaining fine-grained tokens like positional information, adjectives, and colors in multi-turn conversation.}
\label{fig_sup_vis_multi_turn}
\end{figure*}

TAM supports multi-turn conversation for MLLM, which is a new capability compared to conventional models. We first present a qualitative example in Fig. \ref{fig_sup_vis_multi_turn}. Qwen2-VL-2B \cite{wang2024qwen2} can effectively generate the image description, and TAM provides accurate response maps for various attributes, such as objects, actions, and text. Subsequently, the user engaged in multi-turn conversation, inquiring about a fatter dog and the color of a chair. TAM effectively interpreted these fine-grained tokens, including positional information, adjectives, and colors. This example demonstrates TAM's broad applicability and offers strong interpretability analysis for new features like multi-turn dialogue in MLLM.

\begin{figure*}[h]
\centering
 \includegraphics[width=1\textwidth]{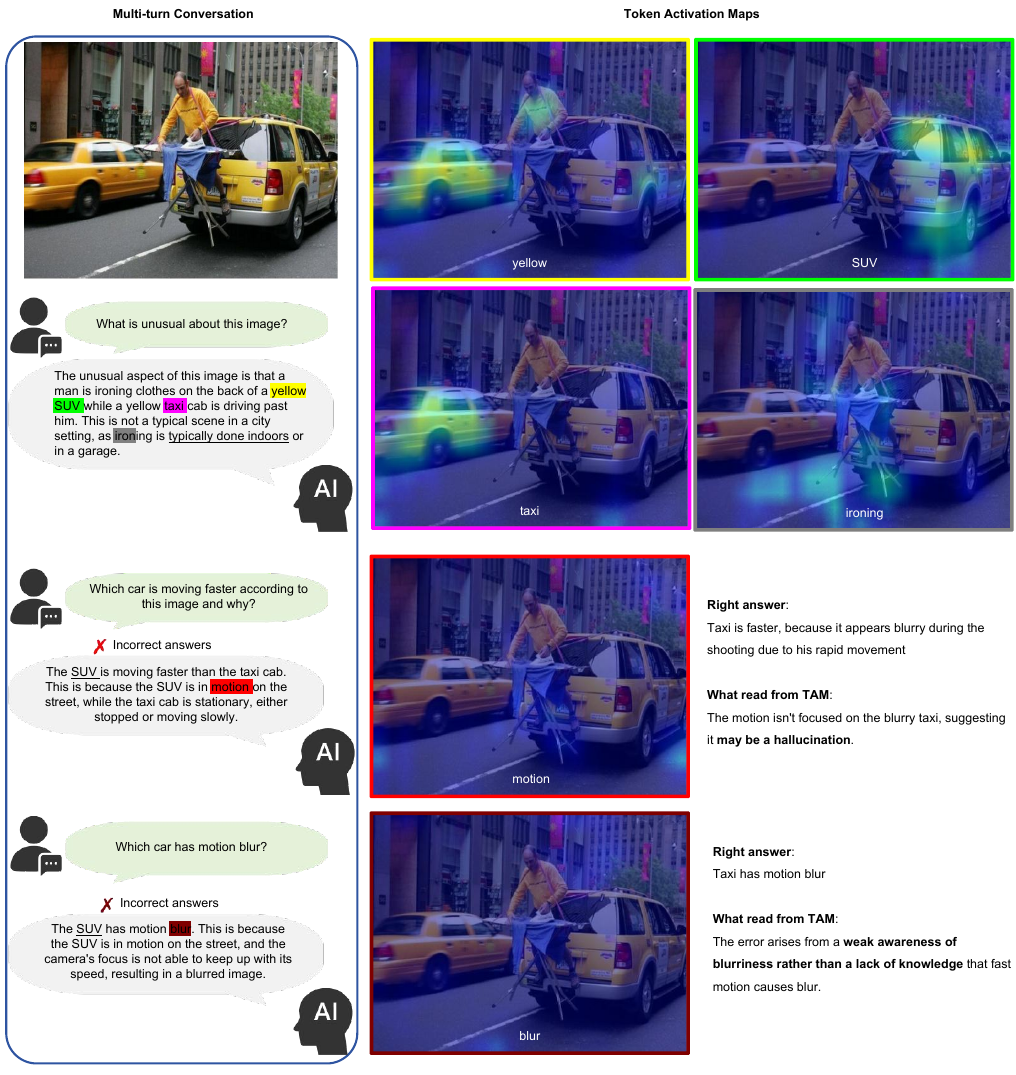}
\caption{\textbf{TAM enables failure case analysis in multi-turn conversation}. Although Qwen2-VL-2B \cite{wang2024qwen2} well recognizes objects with good explanation results in the first round chat, it fails to identify motion blur related to speed and mistakenly regards the SUV as the faster car. The clues provided by TAM reveal that the failure to recognize motion blur is the primary reason for this error, highlighting TAM's effectiveness in supporting detailed error analysis from multi-turn conversation.}
\label{fig_sup_vis_multi_turn2}
\end{figure*}

Additionally, we provide an analysis of a faulty example in Fig. \ref{fig_sup_vis_multi_turn2}, showing that TAM helps locate model errors and provides visual insight for developers. Although Qwen2-VL-2B can recognize why this image is distinctive and demonstrates strong interpretability for object tokens, it made errors in understanding speed and motion blur. Specifically, the taxi exhibited motion blur indicating higher speed, but it incorrectly identified it as an SUV. In the second round of dialogue, we speculated that it might not have recognized the blur, or it could have recognized the blur but failed to understand the relation between blur and speed. Thus, in the third round of dialogue, we asked which vehicle exhibited blur, and the clues provided by TAM indicated that the failure to recognize motion blur was the main reason for this faulty example. TAM supports multi-turn conversation, allowing for more detailed analysis of errors and offering developers precise reasons and a deeper understanding for issues.

\end{document}